\documentclass[11pt]{article}
\pdfoutput=1

\usepackage[margin=1in]{geometry}
\usepackage[utf8]{inputenc}
\usepackage[T1]{fontenc}
\usepackage{hyperref}
\usepackage{url}
\usepackage{booktabs}
\usepackage{amsfonts}
\usepackage{nicefrac}
\usepackage{microtype}
\usepackage{xcolor}
\usepackage{amsmath}
\usepackage{graphicx}
\usepackage{thm-restate}
\usepackage{cleveref}
\usepackage{subcaption}
\usepackage{xfrac}
\usepackage{enumitem}
\usepackage{multirow}
\usepackage{natbib}

\usepackage{amsfonts,bm, amsthm, thmtools}

\newtheorem{theorem}{Theorem}[section]
\newtheorem*{theorem*}{Theorem}
\newtheorem{lemma}[theorem]{Lemma}

\newtheorem{example}[theorem]{Example}

\newtheorem{definition}[theorem]{Definition}

\crefname{observation}{observation}{observations}
\Crefname{observation}{Observation}{Observations}

\def\eqref#1{equation~\ref{#1}}
\def\1{\bm{1}}

\DeclareMathAlphabet{\mathsfit}{\encodingdefault}{\sfdefault}{m}{sl}
\SetMathAlphabet{\mathsfit}{bold}{\encodingdefault}{\sfdefault}{bx}{n}

\newcommand{\E}{\mathbf{E}}

\newcommand{\R}{\mathbb{R}}

\newcommand{\Var}{\mathrm{Var}}

\DeclareMathOperator*{\argmax}{arg\,max}

\title{Truthful Calibration Errors for Multi-Class Prediction}

\author{%
  Yuxuan Lu\thanks{The work was done when Yuxuan Lu was a visiting PhD student at Northwestern University.}\textsuperscript{\, ,}\thanks{Yuxuan Lu and Yifan Wu contributed equally.} \\
  Peking University\\
  \texttt{yx\_lu@pku.edu.cn} \\
  \and
  Yifan Wu\footnotemark[2] \\
  Microsoft Research, New England\\
  \texttt{yifan.wu2357@gmail.com} \\
  \and
  Jason Hartline\\
  Northwestern University\\
  Computer Science\\
  \texttt{hartline@northwestern.edu} \\
  \and
  Lunjia Hu \\
  Northeastern University\\
  Khoury College of Computer Sciences\\
  \texttt{lunjia@alumni.stanford.edu} \\
}
\date{}

\renewcommand{\tilde}{\widetilde}

\newcommand{\truth}{y}
\newcommand{\truthsp}{\mathcal{Y}}

\newcommand{\prob}{p}

\newcommand{\expect}[2]{{\mathbf{E}}_{#1}\left[#2\right]}

\newcommand{\ind}[1]{\mathbb{I}\left[#1\right]}

\newcommand{\pred}{\prob}

\newcommand{\sps}[1]{^{(#1)}}

\newcommand{\CAL}{\textsc{Cal}}

\newcommand{\feature}{x}
\newcommand{\featuresp}{X}

\newcommand{\ndim}{k}

\newcommand{\nsp}{n}

\newcommand{\classifier}{f}

\newcommand{\minspace}[1]{\vspace{#1}}

\begin{document}

\maketitle

\begin{abstract}

Calibrated predictions are useful because their numerical values can be interpreted as probabilities. Calibration errors are therefore widely used to evaluate, compare, and tune probabilistic predictors. Recently, \citet{haghtalab2024truthfulness} introduced an additional requirement for such measures: truthfulness. A calibration measure is truthful if a predictor minimizes its expected measured error by reporting the true conditional label distribution. Many standard empirical calibration errors are non-truthful: a predictor may appear better calibrated by distorting its probabilities rather than reporting them truthfully.

We study the practical role of truthfulness for calibration measurement in multiclass prediction. First, we introduce perfectly truthful calibration errors for multidimensional linear properties of the label distribution, generalizing the truthful calibration error for binary predictions in \citet{hartline2025perfectly}. This framework includes full multiclass calibration and classwise calibration.  We also identify a truthful correction for confidence calibration. Second, we characterize the decision-theoretic implications of these truthful errors. For calibrated predictors, truthful calibration errors preserve the Blackwell dominance: a more informative calibrated predictor receives no larger expected error. Third, we show that this decision-theoretic interpretation explains and mitigates the well-observed ranking robustness problem of binned calibration errors. Empirically, non-truthful confidence-based errors can reverse model rankings when the number of bins changes, while our truthful classwise error gives more stable rankings across binning choices.
\end{abstract}

\minspace{-5mm}
\section{Introduction}
\minspace{-2.5mm}

Probabilistic predictions are useful only when their numerical values can be
interpreted as probabilities.  A weather forecast that reports a $40\%$ chance
of rain should mean that, among days receiving this forecast, rain occurs about
$40\%$ of the time.  This requirement is calibration \citep{dawid}: for a
binary label $y\in\{0,1\}$ and prediction $p\in[0,1]$, calibration asks that $\Pr[y=1\mid p]=p$.

Calibration errors quantify deviations from perfect calibration.  The canonical error is Expected Calibration Error (ECE), which measures the expected
absolute prediction bias: 
$
    \E\!\left[
        \big|p-\Pr[y=1\mid p]\big|
    \right]$.

% This paper asks a basic question: what properties should a calibration error
% have in order to be useful in practice?  
% Calibration errors create incentives. Calibration errors are not used only
% as descriptive statistics.  They are used to compare models, select
% checkpoints, tune post-processing methods, and sometimes train predictors. 
% % Thus a calibration error creates incentives. 
% If a predictor can reduce the
% expected error by distorting its probabilities, then the metric is measuring
% not only calibration, but also the predictor's ability to game the evaluation.
% Recent work
% \citep{haghtalab2024truthfulness,hartline2025perfectly} studies the \textit{truthfulness} of a calibration error. 
% A calibration error is \emph{truthful} if its expected value is minimized by
% reporting the true conditional distribution of the label.

A calibration error creates incentives. If a predictor can reduce its expected measured error by
reporting probabilities different from the true conditional probabilities, then the measure is not only
evaluating calibration; it is also rewarding strategic distortion. \citet{haghtalab2024truthfulness} initiate the study of approximate truthfulness in the online prediction problem. \citet{hartline2025perfectly} propose a perfectly truthful error for binary predictions. In the batch setting, a
calibration measure is truthful if, for every sequence of examples with true conditional label
distributions, the predictor minimizes its expected measured calibration error by reporting these true
conditional distributions. In other words, a truthful calibration error incentivizes true probability
predictions. 

% Truthfulness is the fundamental principle in probabilistic forecasting and
% information elicitation \citep{gneiting2011making,Mcc-56,sav-71}, but its
% practical role for calibration measurement is less understood. Our goal is to
% study what truthfulness implies to calibration errors, especially for the
% multiclass calibration errors commonly used in modern machine learning.  The
% central message is that truthfulness is not merely an abstract incentivization.  For
% the calibration errors we construct, truthfulness gives a decision-theoretic
% interpretation of model rankings: among calibrated predictors, a more
% informative predictor receives a lower expected calibration error.  We
% formalize this statement through Blackwell dominance.

%The need for such a principle already appears in the binary case.  
The following example explains the non-truthfulness of the 
empirical ECE.

\begin{example}[Non-truthfulness of empirical ECE]
\label{example:intro-nontruthful}
Consider $n$ samples whose true label probabilities are
$p_i^*=i/n$, for $i=1,\ldots,n$.  The truthful predictor reports
$p_i=p_i^*$.  If empirical ECE is computed by conditioning on exact prediction
values, then each reported value appears only once, so the empirical estimate
of $\Pr[y=1\mid p_i]$ has constant noise.  The truthful predictor therefore
incurs empirical calibration error of order $O(1)$.  By contrast, a predictor
that always reports $1/2$ pools all samples together and achieves empirical
error of order $O(1/\sqrt n)$.
\end{example}

Binning is the standard practical response to this problem, but remains untruthful.  Instead of
conditioning on exact prediction values, a binned calibration error partitions
the prediction space into $m$ bins and conditions on the bin.   In \Cref{example:intro-nontruthful}, the
truthful predictor has about $n/m$ samples per bin and incurs sampling error of
order $O(\sqrt{m/n})$, while the constant predictor still pools all samples
into one bin and obtains error $O(1/\sqrt n)$.  Thus a non-truthful metric can
prefer a less informative predictor simply because it reports coarser
probabilities.

In this paper, we study the practical role of truthfulness for calibration measurement, especially for
multiclass prediction. Our central message is that truthfulness is not merely an abstract incentive
criterion. For the calibration errors we construct, truthfulness leads to calibration measures with a
decision-theoretic interpretation, and this interpretation helps explain and mitigate the ranking
robustness problems of standard binned calibration errors. 

Our first contribution is to construct truthful calibration errors for multidimensional properties of the
label distribution through minimal changes to canonical binned ECE. The construction replaces the usual $\ell_1$ binned error by its squared, $\ell_2$, version and, for the limiting completeness/soundness guarantee, uses quantile binning. For
linear properties, we prove that this squared binned calibration error is truthful for every
outcome-independent binning rule. The analysis follows the bias--variance decomposition idea in
\citet{hartline2025perfectly}: the expected error separates into a report-dependent squared bias term
and a sampling-variance term, and the bias term is minimized exactly by the true conditional property.
This framework includes full multiclass calibration and classwise calibration as special cases; in our experiments, we use the classwise specialization.

We also examine confidence calibration, a common relaxation of full calibration in the multiclass
setting. Given a multiclass prediction $p \in \Delta(\mathcal{Y})$, the confidence is the largest
reported probability, $\max_{r \in \mathcal{Y}} p_r$. Confidence calibration asks that, among samples
with reported confidence $c$, the predicted top label is correct with probability $c$. We show that
simply applying a truthful binary calibration error to the reported confidence is not truthful. We then
give a corrected confidence calibration error that restores finite-sample truthfulness while preserving
the same asymptotic completeness and soundness requirements of a calibration error.

Our second contribution is to characterize the decision-theoretic implications of these truthful
calibration errors. A downstream user observes a prediction and then chooses an action. If prediction
$f(x)$ Blackwell dominates prediction $g(x)$, then every downstream decision-maker can achieve at
least as much expected utility after observing $f(x)$ as after observing $g(x)$. We show that, in
general, truthful calibration errors preserve this order among calibrated predictors. Specifically for
our constructions, we prove an approximate version showing that the ordering among predictors is
stable when the predictors are approximately calibrated relative to the gap in downstream loss.

Our third contribution is empirical: we show that the above decision-theoretic interpretation helps
mitigate the well-observed ranking robustness problem of binned calibration errors.
%Prior empirical work has observed that model rankings by calibration error may change when the number of bins changes 
Prior empirical work has observed that the relationship between calibration and predictive performance can flip when only the number of bins changes when calculating calibration errors~\citep{nixon2019measuring, minderer2021revisiting}. From our perspective, this instability
is not only a finite-sample artifact: it is a symptom of a non-truthful measure that can reward
non-truthful probabilities. In the approximately calibrated regime identified by our theory, truthful errors preserve
the same rankings of predictors across binning choices up to the remaining calibration bias.
We verify this empirically on neural-network predictors. Non-truthful confidence-based errors can
reverse the ranking between predictors when the number of bins changes, whereas our truthful calibration errors preserve the same qualitative ordering across binning choices.

Together, these results show that truthfulness provides a useful design principle for calibration
measurement and is achievable via simple modifications to existing calibration errors. It aligns the finite-sample evaluation objective with truthful probability reporting,
connects calibration errors to downstream decision quality, and yields more robust empirical model
rankings in multiclass prediction.

\begin{figure}[t]
    \centering
    \begin{subfigure}[t]{0.32\linewidth}
        \centering
        \includegraphics[width=\linewidth]{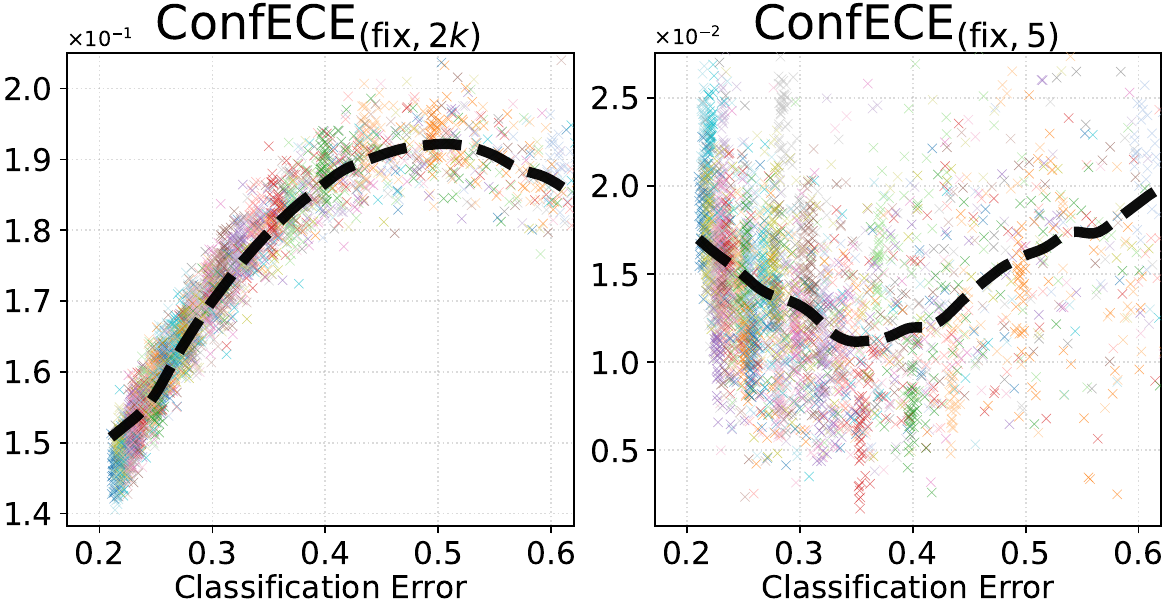}
        \caption{$\mathrm{ConfECE}$ is \textbf{non-truthful}.}
        \label{fig:intro-l1ece-conf}
    \end{subfigure}
    \hfill
    \begin{subfigure}[t]{0.32\linewidth}
        \centering
        \includegraphics[width=\linewidth]{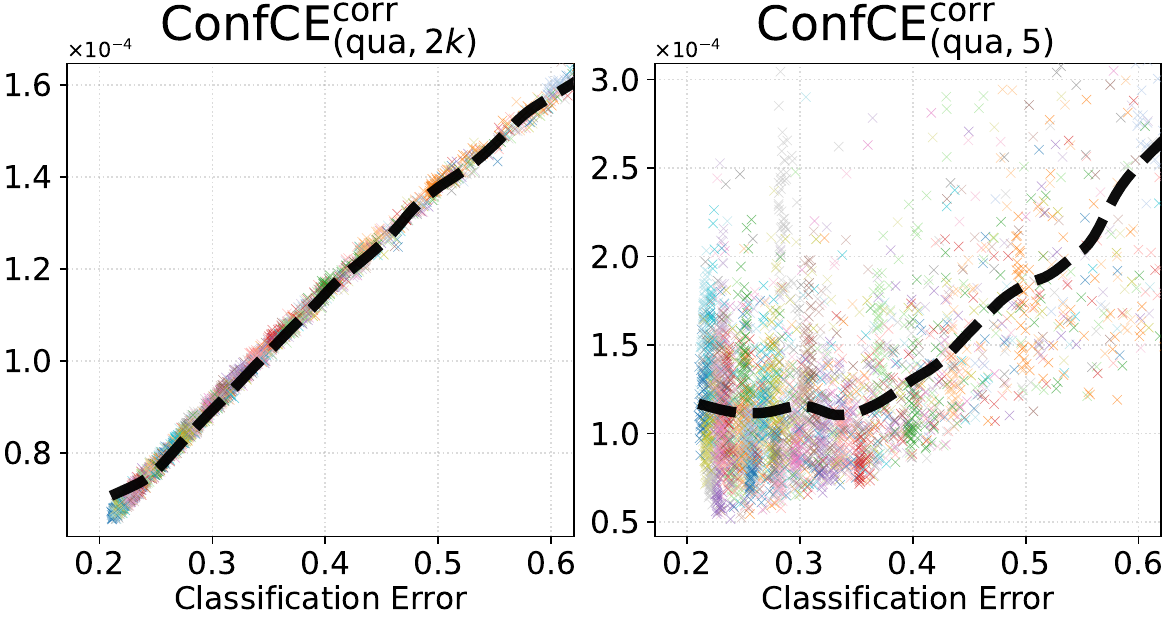}
        \caption{$\mathrm{ConfCE}^{\mathrm{corr}}$ is \textbf{truthful}.}
        \label{fig:intro-l2qece-cconf}
    \end{subfigure}
    \hfill
    \begin{subfigure}[t]{0.32\linewidth}
        \centering
        \includegraphics[width=\linewidth]{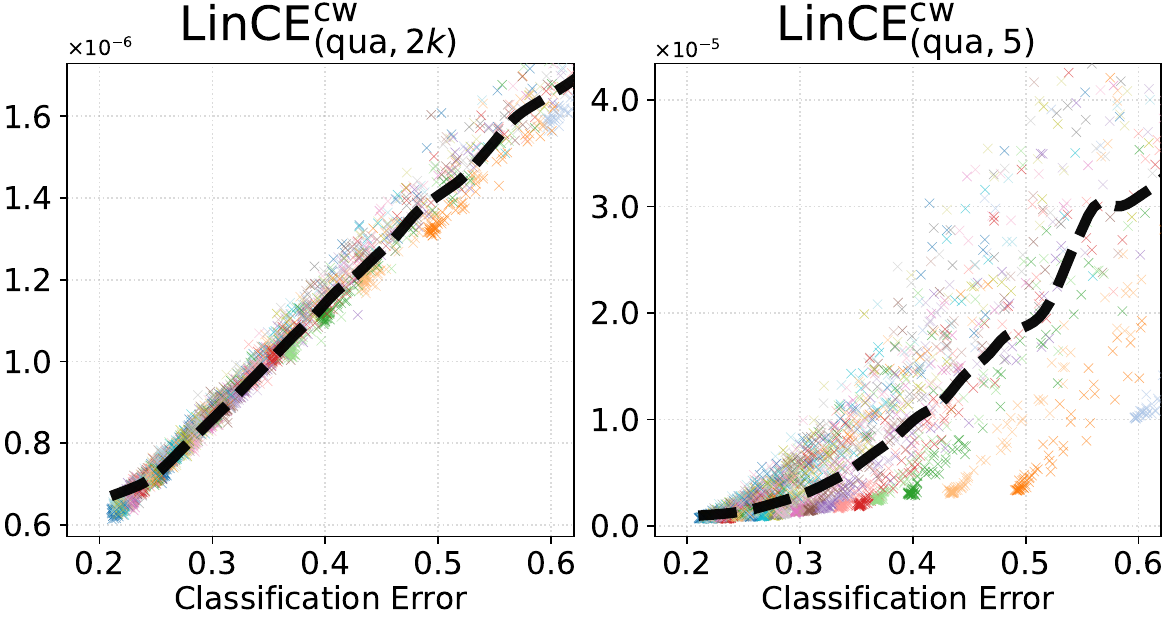}
        \caption{$\mathrm{LinCE}^{\mathrm{cw}}$ is \textbf{truthful}.}
        \label{fig:intro-l2qece-classwise}
    \end{subfigure}
    % \minspace{-2mm}
    \caption{\textbf{Comparison of calibration errors across bin sizes.} Each dot is a classifier checkpoint after temperature scaling on the validation set\protect\footnotemark. Panel (a) shows the canonical confidence-based calibration error used in the literature, namely the same confidence error as in \Cref{example:intro-nontruthful} but estimated after binning predictions into fixed bins. Panel (b) shows our constructed truthful calibration error for confidence, and panel (c) shows our constructed truthful calibration error for classwise calibration. In each panel, the left and right plots use different binning sizes, with $m=2000$ bins and $m=5$ bins, respectively. The qualitative trend is markedly more robust for the truthful errors in (b) and (c): their calibration error remains aligned with predictive quality across bin sizes, whereas the canonical non-truthful error in (a) exhibits a flip in trend when the number of bins changes.
    % Along training, calibration error is expected to first increase and then decrease as loss decreases: at the very beginning, the model predicts almost nothing useful and can appear calibrated; after warmup, its confidence becomes informative but miscalibrated, producing the largest calibration error; with sufficient training, calibration error decreases again as the predictor becomes both accurate and better calibrated.
}

    \label{fig:intro-comparison}
    \minspace{-3mm}
\end{figure}

\footnotetext{
All points are evaluated on the CIFAR-100 test set and come from the same
architecture, MobileNetV3-Small-1.0, trained on different subsets of the
training set.  Points with the same color correspond to checkpoints from the
same training run.
}

    \minspace{-2mm}
\subsection{Related Work}
\minspace{-1mm}
\paragraph{Truthful Calibration Measure.} Recent work introduced the notion of \emph{truthfulness} for a calibration measure. \citet{haghtalab2024truthfulness} first formalize this concept and design an approximately truthful calibration measure in the online setting. \citet{qiao2025truthfulness} study calibration errors that are approximately truthful and quantify decision-making payoff. Both papers focus on approximate truthfulness of a calibration measure in the online setting, while our paper focuses on perfect truthfulness in the batch setting. \citet{hartline2025perfectly} design a perfectly truthful calibration measure for binary classification in the batch setting. Our work generalizes \citet{hartline2025perfectly} beyond binary outcomes to multiclass classification tasks. We also provide empirical evaluations of the truthful error metric. %, where naive reductions to the binary case fail to preserve truthfulness.

    \minspace{-2mm}
\paragraph{Calibration for Multiclass Prediction.}
Calibration in multiclass settings has been studied from several perspectives. \citet{guo2017calibration} introduced confidence calibration methods such as temperature scaling. \citet{minderer2021revisiting} empirically observe that the ranking by calibration error is not robust to hyperparameter (binning size) selection, which motivated our paper. 
%showing that modern neural networks often produce poorly calibrated probabilities and motivating post-hoc adjustments.
\citet{class-wise} discuss the distinction between confidence calibration and classwise calibration. \citet{efficient-multi-class-cal} study projected smooth calibration, reducing a multiclass calibration measure to a binary-class smooth calibration measure. \citet{decision-cal, dim-free-decision} study decision calibration, which calibrates predictions relative to downstream decision tasks. These papers do not study truthfulness, which is the focus of our work.

\paragraph{Non-robustness of Calibration Errors to Bining Granularity.}
It is a widely observed issue that existing binning-based calibration errors are sensitive to the choice of number of bins.
As we mentioned, \citet{minderer2021revisiting} find that different binning granularity can lead to completely opposite conclusions about which models are more calibrated.
The work of \citet{verified} is motivated by recalibration methods that are believed to produce well-calibrated predictions, which turn out to be less calibrated when evaluated using more bins. We show that the truthful calibration errors we introduce gives a solution to this issue, achieving robustness and consistency accross different choices of binning granularity.

\minspace{-2mm}
\section{Preliminaries}
\label{sec:prelim}
\minspace{-2mm}

Each sample $(x,y)\sim D$
consists of a feature $x\in\featuresp$ and an outcome $y\in\truthsp$.
For multiclass prediction, $\truthsp=\{1,\ldots,k\}$.  We write
$[n]=\{1,\ldots,n\}$ and let $\ind{E}$ denote the indicator of event $E$.
Throughout, every $\argmax$ uses a fixed deterministic tie-breaking rule.

\paragraph{Linear properties.}
A \emph{linear property} is specified by a map
$\phi:\truthsp\to\R^d$ and maps a distribution $P\in\Delta(\truthsp)$ to
\[
    \Gamma_\phi(P) := \E_{y\sim P}[\phi(y)]
    \in \mathcal R_\phi,
    \qquad
    \mathcal R_\phi := \operatorname{conv}\{\phi(y):y\in\truthsp\}.
\]
A predictor for $\Gamma_\phi$ is a map
$g:\featuresp\to\mathcal R_\phi$.  It is calibrated for $\Gamma_\phi$ if
\[
    \E[\phi(y)\mid g(x)=u] = u
    \qquad
    \text{for every report }u\in\mathcal R_\phi .
\]
Binary calibration is the special case $\phi(y)=y$ for
$y\in\{0,1\}$.  Full multiclass calibration is the special case
$\phi(y)=e_y$, where $e_y$ is the $y$-th standard basis vector.
Classwise calibration corresponds to the $k$ scalar properties
$\phi_r(y)=\ind{y=r}$, evaluated on the coordinate reports $p_r$.

\paragraph{Truthfulness.}
A batch calibration measure for $\Gamma_\phi$ maps
$n$ reports and outcomes to a real number:
\[
    \CAL_\phi:\mathcal R_\phi^n\times\truthsp^n\to\R .
\]
It is \emph{truthful} if reporting the true property values minimizes
expected error.  Formally, for any distributions
$P_1,\ldots,P_n\in\Delta(\truthsp)$, let
\[
    \mu_i^* := \Gamma_\phi(P_i)=\E_{y_i\sim P_i}[\phi(y_i)] .
\]
Then $\CAL_\phi$ is truthful if, for every alternative report
$\mu_1,\ldots,\mu_n\in\mathcal R_\phi$,
\[
    \E\big[\CAL_\phi(\mu^*_{1:n};y_{1:n})\big]
    \le
    \E\big[\CAL_\phi(\mu_{1:n};y_{1:n})\big],
\]
where $y_i\sim P_i$ independently.

\paragraph{Completeness and soundness.}
Completeness and soundness are different from truthfulness: they are asymptotic
properties that detect miscalibration, not finite-sample reporting incentives.
Let $\{\CAL_n\}_{n\ge 1}$ be a sequence of sample calibration measures for a
calibration notion $\mathsf C$; for example, $\mathsf C$ may be calibration for
a linear property $\Gamma_\phi$, classwise calibration, or confidence
calibration.  We say that the sequence $\{\CAL_n\}_{n\ge 1}$ is
\emph{complete in the limit} for $\mathsf C$ if, for every
$\mathsf C$-calibrated predictor $h$,
\[
    \lim_{n\to\infty}
    \E\!\left[
        \CAL_n(h(x_1),\ldots,h(x_n);y_1,\ldots,y_n)
    \right]
    =
    0 .
\]
It is \emph{sound in the limit} for $\mathsf C$ if, for every predictor $h$
that is not $\mathsf C$-calibrated,
\[
    \liminf_{n\to\infty}
    \E\!\left[
        \CAL_n(h(x_1),\ldots,h(x_n);y_1,\ldots,y_n)
    \right]
    >
    0 .
\]
For binned measures, the limit is taken along the specified sequence of
binning rules.  Deterministic positive rescalings of $\CAL_n$ do not affect
truthfulness or predictor rankings, but they can be useful when matching a
finite-sample error to its population calibration target.

\paragraph{Binning.}
A binning rule $\mathsf B$ maps reports $u_{1:n}$ to a partition
$\mathsf B(u_{1:n})$ of $[n]$ and is not allowed to depend on the
realized outcomes.  For scalar reports, two standard choices are fixed
bins and quantile bins.  Fixed bins first choose intervals
$I_1,\ldots,I_m$ that partition the report space and then set
$B_j=\{i\in[n]:u_i\in I_j\}$.  In the common case of reports in
$[0,1]$, one may take equal-width bins
$I_j=((j-1)/m,j/m]$ for $j=1,\ldots,m$, with $0\in I_1$.  Quantile
binning with $m$ bins sorts the reports and sets
\[
    B_j
    =
    \left\{
    i\in[n]\;:\;
    \frac{(j-1)n}{m}< i \le \frac{jn}{m}
    \right\},
    \qquad j=1,\ldots,m .
\]
For a sequence $m_n$, we call the
associated sequence of quantile binning rules $\{\mathsf B_{m_n}^Q\}_{n\ge1}$
a \emph{consistent regime} if $m_n\to\infty$ and $n/m_n\to\infty$, i.e., the number of bins grows while the number of samples per
bin also diverges.  The truthfulness results below hold for any
outcome-independent binning rule.

\begin{definition}[Binned calibration error for a linear property]
\label{def:lince}
For a map $\phi:\truthsp\to\R^d$ and a binning rule $\mathsf B$, define the $\ell_1$ and $\ell_2$ version of calibration error:
\begin{align*}
    \mathrm{LinECE}_{\phi,\mathsf B}
    (u_{1:n};y_{1:n})
    :=&
    \frac{1}{n}
    \sum_{B\in\mathsf B(u_{1:n})}
    \left\|
        \sum_{i\in B}\bigl(u_i-\phi(y_i)\bigr)
    \right\|_1,\\
    \mathrm{LinCE}_{\phi,\mathsf B}
    (u_{1:n};y_{1:n})
    :=&
    \frac{1}{n^2}
    \sum_{B\in\mathsf B(u_{1:n})}
    \left\|
        \sum_{i\in B}\bigl(u_i-\phi(y_i)\bigr)
    \right\|_2^2 .
\end{align*}
\end{definition}

When $d=1$, $\phi(y)=y$, and $\mathsf B$ is quantile binning,
$\mathrm{LinCE}_{\phi,\mathsf B}$ is the squared quantile-binned
calibration error $\ell_2$-QECE of \citet{hartline2025perfectly}, while
$\mathrm{LinECE}_{\phi,\mathsf B}$ is its $\ell_1$ analogue.  Our
truthfulness results below use the squared $\ell_2$ form
$\mathrm{LinCE}_{\phi,\mathsf B}$.

\minspace{-2mm}
\section{Truthful Calibration Measures}
\label{sec:truthful}
\minspace{-2mm}

The key observation is that squared binned residuals admit a
bias-variance decomposition for every linear property.  This gives a
single truthfulness theorem that covers binary calibration, full
multiclass calibration, classwise calibration, and other moment properties of the outcome.

\subsection{Truthful Calibration Error for Linear Properties}

\begin{theorem}[Truthfulness for linear properties]
\label{thm:linear-property-truthful}
For every signal $\phi:\truthsp\to\R^d$ and every outcome-independent
binning rule $\mathsf B$, the measure
$\mathrm{LinCE}_{\phi,\mathsf B}$ is truthful for the linear property
$\Gamma_\phi(P)=\E_{P}[\phi(y)]$.

More precisely, let $P_1,\ldots,P_n\in\Delta(\truthsp)$ be arbitrary,
let $\mu_i^*=\Gamma_\phi(P_i)$, and draw
$y_i\sim P_i$ independently.  Then for every report
$u_{1:n}\in\mathcal R_\phi^n$,
\[
\begin{aligned}
    \E\big[
    \mathrm{LinCE}_{\phi,\mathsf B}
    (u_{1:n};y_{1:n})
    \big]
    &=
    \frac{1}{n^2}
    \sum_{B\in\mathsf B(u_{1:n})}
    \left\|
        \sum_{i\in B}(u_i-\mu_i^*)
    \right\|_2^2  +
    \frac{1}{n^2}
    \sum_{i=1}^n
    \E_{y_i\sim P_i}
    \left[
        \|\phi(y_i)-\mu_i^*\|_2^2
    \right].
\end{aligned}
\]
Thus the truthful report $u_i=\mu_i^*$ makes the bias term vanish and
minimizes the expected error.  Moreover, in the scalar case $d=1$, if
the binning rule is quantile binning in the consistent regime, then the sequence
$\{\mathrm{LinCE}_{\phi,\mathsf B_n}\}_{n\ge1}$ is complete and sound
in the limit for calibration of $\Gamma_\phi$.
\end{theorem}
The proof is deferred to \Cref{apdx:truthful}.
% \begin{proof}
% Fix the reports $u_{1:n}$.  The bins are then fixed independently of
% the realized states.  For any bin $B$,
% \[
%     \E\left[
%     \left\|
%         \sum_{i\in B}(u_i-\phi(y_i))
%     \right\|_2^2
%     \right]
%     =
%     \left\|
%         \sum_{i\in B}(u_i-\mu_i^*)
%     \right\|_2^2
%     +
%     \sum_{i\in B}
%     \E\left[
%         \|\phi(y_i)-\mu_i^*\|_2^2
%     \right],
% \]
% where the cross terms vanish by independence and
% $\E[\phi(y_i)]=\mu_i^*$.  Summing over bins gives the displayed
% identity.  The second term is independent of the report, while the first
% term is nonnegative and equals zero at the truthful report.
% \end{proof}
Since the truthful value is only the sampling-variance term, it is
$O(1/n)$ for bounded $\phi$.  With the usual consistent binning regime
in which bins shrink while their sample sizes grow, the remaining bias
term gives the population calibration error for the property
$\Gamma_\phi$.

\paragraph{Classwise calibration as a linear property.}
For multiclass prediction, take
$\phi_r(y)=\ind{y=r}$ for each class $r$.  Given probabilistic reports
$p_i\in\Delta(\truthsp)$, define the classwise squared binned error by
running the scalar rule separately on each coordinate with a scalar
outcome-independent binning rule $\mathsf B$:

\minspace{-3mm}
\[
    \mathrm{LinCE}^{\mathrm{cw}}_{\mathsf B}
    (p_{1:n};y_{1:n})
    :=
    \frac{1}{k}
    \sum_{r=1}^k
    \mathrm{LinCE}_{\phi_r,\mathsf B}
    \bigl(p^{(r)}_{1:n};y_{1:n}\bigr),
\]

\minspace{-3mm}

where $p_i^{(r)}$ is the $r$-th coordinate of $p_i$ and
$\mathsf B$ is applied to the scalar reports
$p_1^{(r)},\ldots,p_n^{(r)}$ for each class $r$.  By
\Cref{thm:linear-property-truthful}, this measure is truthful for
classwise calibration.  Equivalently, truthfulness is preserved by
averaging the truthful scalar errors for the $k$ one-vs-all linear
properties.  The same vector construction also covers full multiclass
calibration through $\phi(y)=e_y$; in the experiments below, we focus on
the classwise specialization.

\minspace{-2mm}
\subsection{Confidence Calibration}
\label{sec:confidence-calibration}
\minspace{-2mm}

Confidence calibration differs from the linear properties above because
the evaluated coordinate depends on the report itself.  For a multiclass
report 
$p_i\in\Delta(\truthsp)$, let
\minspace{-2mm}
\[
    r_i := \argmax_{r\in\truthsp} p_i^{(r)},\qquad
    c_i := p_i^{(r_i)},\qquad
    z_i := \ind{y_i=r_i}.
\]
The usual confidence aggregation applies a binary calibration error to
the pairs $(c_i,z_i)$, which is not generally truthful: a predictor can
change which class becomes the argmax and thereby change the binary
problem being evaluated.  A concrete counterexample is given in
\Cref{apdx:confidence-nontruthful-example}. %: even when the underlying
% binary calibration error is truthful, confidence aggregation is non-truthful.

The following corrected confidence measure adds a vanishing
classification-error term $\frac{1}{n}\left[1-\frac{1}{n}\sum_{i=1}^n z_i\right]$.  The correction is at most $1/n$, so it does not
change the limiting confidence-calibration target, but it does restore
finite-sample truthfulness.

\begin{definition}[Corrected squared confidence error]
\label{def:corrected-confidence}
Let $\mathsf B$ be a binning rule applied to the scalar confidences
$c_{1:n}$.  Define
\minspace{-2mm}
\[
\begin{aligned}
    \mathrm{ConfCE}^{\mathrm{corr}}_{\mathsf B}
    (p_{1:n};y_{1:n})
    &:=
    \frac{1}{n^2}
    \sum_{B\in\mathsf B(c_{1:n})}
    \left(
        \sum_{i\in B}(c_i-z_i)
    \right)^2
    +
    \frac{1}{n}
    \left[
        1-\frac{1}{n}\sum_{i=1}^n z_i
    \right].
\end{aligned}
\]
\minspace{-2mm}
\end{definition}

\begin{theorem}[Truthful confidence calibration]
\label{thm:confidence-truthful}
For every outcome-independent binning rule $\mathsf B$,
$\mathrm{ConfCE}^{\mathrm{corr}}_{\mathsf B}$ is truthful for confidence
calibration. If
the binning rule is quantile binning in the consistent regime, then the sequence
$\{\mathrm{ConfCE}^{\mathrm{corr}}_{\mathsf B_n}\}_{n\ge1}$ is complete and
sound in the limit for confidence calibration.

% More explicitly, let $p_1^*,\ldots,p_n^*\in\Delta(\truthsp)$ be the
% true class distributions and draw $y_i\sim p_i^*$ independently.  For
% any reported distributions $p_1,\ldots,p_n$, let
% $r_i=\argmax_r p_i^{(r)}$ and $c_i=p_i^{(r_i)}$.  Then
% \[
% \begin{aligned}
%     n^2\,
%     \E\big[
%     \mathrm{ConfCE}^{\mathrm{corr}}_{\mathsf B}
%     (p_{1:n};y_{1:n})
%     \big]
%     &=
%     \sum_{B\in\mathsf B(c_{1:n})}
%     \left(
%         \sum_{i\in B}
%         \bigl(c_i-(p_i^*)^{(r_i)}\bigr)
%     \right)^2  \\
%     &\qquad
%     +
%     \sum_{i=1}^n
%     \left(
%         1-\bigl((p_i^*)^{(r_i)}\bigr)^2
%     \right).
% \end{aligned}
% \]
% Hence the expected error is minimized by selecting a class
% $r_i\in\argmax_r (p_i^*)^{(r)}$ and reporting confidence
% $c_i=(p_i^*)^{(r_i)}$.  In particular, the full truthful report
% $p_i=p_i^*$ is optimal.
\end{theorem}
The proof is deferred to \Cref{apdx:truthful}.

% \begin{proof}
% Fix the reported distributions, and write
% $q_i=(p_i^*)^{(r_i)}$.  Then $a_i=\ind{y_i=r_i}$ is an independent
% Bernoulli random variable with mean $q_i$.  The squared confidence term
% satisfies
% \[
%     \E\left[
%     \sum_{B\in\mathsf B(c_{1:n})}
%     \left(
%         \sum_{i\in B}(c_i-z_i)
%     \right)^2
%     \right]
%     =
%     \sum_{B\in\mathsf B(c_{1:n})}
%     \left(
%         \sum_{i\in B}(c_i-q_i)
%     \right)^2
%     +
%     \sum_{i=1}^n q_i(1-q_i).
% \]
% The correction contributes
% \[
%     n^2\,
%     \E\left[
%     \frac{1}{n}
%     \left(
%         1-\frac{1}{n}\sum_i a_i
%     \right)
%     \right]
%     =
%     n-\sum_{i=1}^n q_i .
% \]
% Adding the two displays gives
% \[
%     n^2\E[\mathrm{ConfCE}^{\mathrm{corr}}_{\mathsf B}]
%     =
%     \sum_{B\in\mathsf B(c_{1:n})}
%     \left(
%         \sum_{i\in B}(c_i-q_i)
%     \right)^2
%     +
%     \sum_{i=1}^n (1-q_i^2).
% \]
% The first term is nonnegative and is zero when $c_i=q_i$.  The second
% term is minimized by choosing $q_i$ as large as possible, i.e., by
% selecting a true most-likely class.  Thus truthful confidence reporting
% minimizes expected error.
% \end{proof}

\minspace{-2mm}
\paragraph{Why the correction is necessary.}
Without the correction, the expected confidence error contains the
variance term $\sum_i q_i(1-q_i)$ rather than
$\sum_i(1-q_i^2)$.  The former is not monotone in the selected true
probability $q_i$, so changing the reported argmax can reduce the
measured error even when the confidence report is not truthful.  The
correction changes the objective so that selecting the true most-likely
class and reporting its probability is optimal.

\minspace{-2mm}
\section{Blackwell Dominance of the Constructed Truthful Errors}
\label{sec:blackwell}
\minspace{-2mm}

Calibration errors quantify different kinds of miscalibration.  Linear-property
errors test whether the reported expectation of $\phi(y)$ is conditionally
correct; classwise errors test this one label at a time; and confidence errors
test the reported probability of the predicted top label.  These choices can
lead to different numerical values.  Truthfulness imposes a more basic ranking
principle: among calibrated predictors, a more informative prediction should
not receive a larger expected error.

The first two results below use only the definition of truthfulness.  They show
that any truthful batch calibration measure preserves Blackwell dominance among
calibrated predictors, and that canonical recalibration can only decrease the
expected truthful error.  We then instantiate these conclusions for the squared-
residual errors constructed in this paper.  For
$\mathrm{LinCE}_{\phi,\mathsf B}$ and its classwise specialization, the linear-
property theorems below apply directly.  For
$\mathrm{ConfCE}^{\mathrm{corr}}_{\mathsf B}$, confidence calibration is report-
dependent rather than a linear property, so its ranking interpretation will
come from the proper-loss identity in \Cref{sec:approx-blackwell}.

\paragraph{Blackwell dominance between predictions.}

A downstream user acts after seeing a prediction.  For a predictor
$h:\featuresp\to\mathcal Z_h$, define
\[
    \pi_h(z) := \Pr[y\mid h(x)=z]
\]
for the posterior distribution over outcomes after observing the prediction
$z$. For calibrated predictors, $\pi_h(z)$ is the identity function.  We say that prediction $f$ \emph{Blackwell dominates} prediction $g$,
written $f\succeq_B g$, if there exists a Markov kernel $K$ such that $g(x)$
can be simulated from $f(x)$ via $K$ without observing $y$ \citep{blackwell1951comparison}.  Equivalently, for
every bounded downstream decision problem $u:A\times\truthsp\to[0,1]$,
\[
    \E\left[
        \max_{a\in A}
        \E[u(a,y)\mid f(x)]
    \right]
    \ge
    \E\left[
        \max_{a\in A}
        \E[u(a,y)\mid g(x)]
    \right].
\]
In words, after best-responding to the observed prediction, a decision-maker
who sees $f(x)$ can achieve at least as much expected utility as one who sees
$g(x)$, in every downstream task.
%A downstream user acts by selecting decision $a\in A$ and gaining utility  $u:A\times y\to[0,1]$ after seeing a prediction.  For a predictor
% $h:\featuresp\to\mathcal Z_h$, define the best-response payoff of a decision maker who takes best response $a^*$: 
% \[
%     \ell_u(h(x), y) = u(a^* (h(x)), y), \text{ where }a^*(h(x)) = \argmax_{a\in A}\E_{y\sim h(x)}\left[u(a, y)\right]. 
% \]
% % for the posterior distribution over outcomes after observing the prediction
% $z$.  
% We say that prediction $f$ \emph{Blackwell dominates} prediction $g$,
% written $f\succeq_B g$, if 
% there exists a Markov kernel $K$ such that $g(x)$
% can be simulated from $f(x)$ via $K$ without observing $y$.  Equivalently, 
% for
% every bounded downstream decision problem $u:A\times\truthsp\to[0,1]$,
% \[
%     \E\left[
%        \ell_u(f(x), y) \right]
%     \ge
%     \E\left[
%         \ell_u(h(x), y)
%     \right].
% \]
% In words, after best-responding to the observed prediction, a decision-maker
% who sees $f(x)$ can achieve at least as much expected utility as one who sees
% $g(x)$, in every downstream task.

\subsection{Consequence of Truthfulness: Preserving Orders between Calibrated Predictors}
This section presents the batch analogue of the standard fact that proper losses
preserve Blackwell dominance for single predictions.  The main difference of a truthful calibration error from a proper loss is the composition of multiple independent samples as the input to the error. \Cref{thm:blackwell-general} shows that the Blackwell ordering among calibrated predictors generalizes under independent composition of samples. 

\begin{theorem}[Truthful batch measures preserve Blackwell dominance]
\label{thm:blackwell-general}
Let $\CAL_\phi:\mathcal R_\phi^n\times\truthsp^n\to\R$ be truthful for the
linear property $\Gamma_\phi$.  Let
$f,g:\featuresp\to\mathcal R_\phi$ be calibrated for $\Gamma_\phi$.  If
$f\succeq_B g$, then
\[
    \E\!\left[
        \CAL_\phi(f(x_{1:n});y_{1:n})
    \right]
    \le
    \E\!\left[
        \CAL_\phi(g(x_{1:n});y_{1:n})
    \right],
\]
where $(x_i,y_i)_{i=1}^n$ are drawn i.i.d.\ from $D$.
\end{theorem}

\Cref{thm:recalibration-general} shows that truthful  calibration errors reward
recalibration. 

\begin{theorem}[Canonical recalibration reduces expected truthful error]
\label{thm:recalibration-general}
Let $\CAL_\phi:\mathcal R_\phi^n\times\truthsp^n\to\R$ be truthful for the
linear property $\Gamma_\phi$, and let
$h:\featuresp\to\mathcal R_\phi$ be any predictor.  Define its canonical
recalibration by
\[
    \bar h(x)
    :=
    \E[\phi(y)\mid h(x)].
\]
Then
\[
    \E\!\left[
        \CAL_\phi(\bar h(x_{1:n});y_{1:n})
    \right]
    \le
    \E\!\left[
        \CAL_\phi(h(x_{1:n});y_{1:n})
    \right].
\]
\end{theorem}

Proofs of \Cref{thm:blackwell-general,thm:recalibration-general} are deferred
to \Cref{apdx:general-blackwell}.

\minspace{-2mm}
\subsection{The Order between Approximately Calibrated Predictors}
\label{sec:approx-blackwell}
\minspace{-2mm}

The previous theorem gives the ranking for calibrated predictors.  For
approximately calibrated predictors, the same order remains stable when the
proper-loss gap dominates the calibration bias.  We note that the theorems in this section are
about the constructed errors above, not about all truthful calibration
measures. 

Let $\CAL$ denote one of the constructed errors:
$\mathrm{LinCE}_{\phi,\mathsf B}$,
$\mathrm{LinCE}^{\mathrm{cw}}_{\mathsf B}$, or
$\mathrm{ConfCE}^{\mathrm{corr}}_{\mathsf B}$.  Define the associated losses by
\minspace{-3mm}
\begin{align*}
    \ell_\phi(u,y)&:=\|u-\phi(y)\|_2^2, \quad
    \ell_{\mathrm{Brier}}(p,y)
    :=
    \sum_{r=1}^k \bigl(p^{(r)}-\ind{y=r}\bigr)^2,\\
    \ell_{\mathrm{conf}}(p,y)
    &:=
    1-\ind{y=r_p}
    +
    \bigl(p^{(r_p)}-\ind{y=r_p}\bigr)^2,
    \qquad
    r_p\in\argmax_{r\in\truthsp} p^{(r)}.
\end{align*}
\minspace{-3mm}

When the predictor is calibrated for the relevant notion, the expected
calibration error equals the associated proper-loss risk, summarized in the
following table.  The confidence row is not a specialization of the linear-
property theorem above; rather, it follows from the separate proper-loss
identity for $\mathrm{ConfCE}^{\mathrm{corr}}_{\mathsf B}$ proved in
\Cref{apdx:general-blackwell}.
\[
\begin{array}{c|c}
\text{constructed error } \CAL & \text{associated loss } \ell_\CAL \\ \hline
\mathrm{LinCE}_{\phi,\mathsf B}
&
\ell_\phi(u,y)=\|u-\phi(y)\|_2^2
\\[2mm]
\mathrm{LinCE}^{\mathrm{cw}}_{\mathsf B}
&
\ell_{\mathrm{Brier}}(p,y)/k
\\[2mm]
\mathrm{ConfCE}^{\mathrm{corr}}_{\mathsf B}
&
\ell_{\mathrm{conf}}(p,y)
\end{array}
\]
\minspace{-3mm}

% \yuxuan{notation}

Any of these losses may be rescaled by a deterministic positive constant,
if desired, without affecting the ordering statements below.  For a possibly
miscalibrated predictor $h$, let $\bar h$ be its canonical recalibration: it
keeps the same prediction $h(x)$ but replaces the report by the true
conditional property given that prediction.  Thus, for a linear property,
\[
    \bar h_\phi(x)
    :=
    \E[\phi(y)\mid h(x)],
\]
and for full multiclass prediction,
$
    \bar h(x)
    :=
    \pi_h(h(x))$.

For each constructed error, define the calibration-bias term
$\rho_\CAL(h)\ge 0$ by the decomposition
\[
    n\,\E[\CAL(h)]
    =
    \E\!\left[
        \ell_\CAL(\bar h(x),y)
    \right]
    +
    \rho_\CAL(h).
\]
For $\mathrm{LinCE}_{\phi,\mathsf B}$, this bias is the squared binned residual
between the reported property $h(x)$ and the canonical property
$\bar h_\phi(x)$.  For corrected confidence calibration, it includes both
confidence miscalibration and the regret from selecting a label that is not
most likely under the posterior.

\begin{theorem}[Approximate calibrated orders for the constructed errors]
\label{thm:approx-constructed-order}
Fix one of the constructed errors
$\CAL\in\{\mathrm{LinCE}_{\phi,\mathsf B},
\mathrm{LinCE}^{\mathrm{cw}}_{\mathsf B},
\mathrm{ConfCE}^{\mathrm{corr}}_{\mathsf B}\}$,
and let $\ell_\CAL$ be its associated proper loss.  Suppose predictors
$f$ and $g$ satisfy
\[
    \rho_\CAL(f)\le \epsilon .
\]
If their canonically recalibrated proper-loss risks are separated by
\[
    \E\!\left[
        \ell_\CAL(\bar f(x),y)
    \right]
    +
    \epsilon
    \le
    \E\!\left[
        \ell_\CAL(\bar g(x),y)
    \right],
\]
then the constructed calibration error preserves the ranking:
\[
    \E[\CAL(f)]
    \le
    \E[\CAL(g)].
\]

Equivalently, in payoff form, if the bounded scoring payoff associated with
$\ell_\CAL$ favors $\bar f$ over $\bar g$ by at least $\epsilon$, then the
constructed truthful error preserves the ranking.
\end{theorem}
The proof is deferred to \Cref{apdx:general-blackwell}.  Equivalently, the
associated bounded scoring task favors $\bar f$ over $\bar g$ by the same
margin.

    \minspace{-3mm}
\section{Empirical Evaluations}
\label{sec: empirical}
    \minspace{-2mm}
% In this section, we conduct empirical evaluations of neural network predictors with different calibration measures. We study the errors' robustness to binning size selection as an implication of the dominance-preserving property. 
% When a truthful calibration measure preserves the dominance, the ranking between (approximately) calibrated predictors is consistent across different binning size selections. %When the predictors are sufficiently close to calibration, the truthful calibration measure remains robust to the selection of a binning size. 

Our experiments follow the standard post-hoc calibration pipeline used in
empirical studies of neural-network calibration, such as 
\citet{minderer2021revisiting}.  We first train neural-network classifiers on
standard multiclass prediction tasks, then apply simple post-hoc calibration
methods to their predicted probabilities.  %This separates two effects: the predictive information
% contained in the neural network, and the calibration quality of the reported
% probabilities after post-hoc adjustment.

\paragraph{Experimental goal.} The purpose of the experiments is not to propose a new calibration algorithm, but to validate the assumptions and implications of our theory in a realistic
neural-network setting, and to clarify the source of the inconsistent rankings
observed in prior work.  Different calibration errors, and even the same error
with different hyperparameters, such as the number of bins, naturally quantify
calibration bias in different ways.  We therefore should not expect all errors
to produce identical numerical values.  However, they should still respect
certain fundamental decision-theoretic principles: among approximately
calibrated predictors, the ranking induced by a calibration error should agree
with their Blackwell ordering, up to the
remaining calibration bias. 

Our theoretical results formalize this principle.  They assume that the
predictors are approximately calibrated and show that small remaining
calibration errors should not overturn the ranking induced by the associated
bounded scoring losses.  The experiments test this theory.  We observe that,
first, after post-hoc calibration, neural networks are approximately calibrated;
second, the constructed truthful calibration errors preserve the same ranking as
the corresponding bounded scoring losses, invariant to binning sizes.  This also
helps reinterpret prior empirical observations: the inconsistent rankings found
there arise precisely in settings where calibration errors violate these
decision-theoretic principles.  Thus, our experiments both replicate the
standard post-hoc calibration pipeline and support the theoretical observation
that approximate calibration is sufficient for ranking preservation.
% : once the calibration bias is small, the
% truthful calibration error reflects downstream decision quality rather than
% artifacts of miscalibration.

    \minspace{-2mm}
\subsection{Experimental Setup}
    \minspace{-2.5mm}
% \textbf{Dataset. } We use dataset \texttt{CIFAR-100}. \lunjia{Should we also describe how we split the dataset for training / validation /test?}

% \textbf{Model. } We use MobileNetV3-Small, ResNet10/18/34/50. We keep each checkpoint in the training process. We evaluate the model after temperature scaling as in \citet{minderer2021revisiting}. For temperature scaling, we select the temperature of the model output to minimize the log loss (a.k.a., cross-entropy loss) on the validation set. 

\textbf{Dataset. } We use the \texttt{CIFAR-100} dataset, split into 50000 training images, 5000 validation images, and 5000 test images~\citep{krizhevsky2009learning}. The validation split is used only for temperature scaling, and all reported empirical results are evaluated on the test split.

\textbf{Models. } We evaluate nine pretrained image classifiers fine-tuned on \texttt{CIFAR-100}: MobileNetV3-Small-0.5/0.75/1.0 and ResNet10t/18/34/50/101/152. For each training run, we keep model checkpoints throughout fine-tuning and treat them as a sequence of probabilistic predictors along training. Following \citet{minderer2021revisiting}, we apply temperature scaling to every checkpoint, with the temperature parameter chosen to minimize the cross-entropy on the validation set.

\textbf{Evaluation pool. } Our experiments use two complementary collections of predictor traces. For the within-model experiments, we focus on MobileNetV3-Small-1.0 and collect 40 training traces obtained from random training subsets of different sizes. This provides a dense view of how calibration errors behave across checkpoints within a single architecture. For the cross-model experiments, we consider all nine architectures and use 5 training traces for each model, again obtained from random training subsets of different sizes. This yields a diverse pool of predictors for comparing calibration rankings across architectures.

\textbf{Binning. } Empirically, we find that fixed-bin and quantile-binned evaluations give very similar trends in our experiments. Their main difference is therefore theoretical rather than empirical: truthfulness holds for any outcome-independent binning rule, while quantile binning is used for the limiting completeness/soundness guarantee and for the main empirical presentation. For this reason, we use quantile-binned calibration errors in the main text and defer most fixed-bin results to the appendix. When we need to specify the binning rule and the number of bins explicitly for any calibration error $\CAL$, we write $\CAL_{(\mathrm{qua},m)}$ for the version with $m$ quantile bins and $\CAL_{(\mathrm{fix},m)}$ for the version with $m$ fixed bins.

\textbf{Evaluation metrics. } We evaluate predictors using several losses, including classification error, cross-entropy loss, Brier loss, spherical loss, and confidence loss. Across our model checkpoints, these losses exhibit nearly linear trends with one another, so for simplicity we use classification error as the primary reference loss. We compare our truthful $\mathrm{LinCE}^{\mathrm{cw}}_{\mathsf B}$ (classwise calibration) and $\mathrm{ConfCE}^{\mathrm{corr}}_{\mathsf B}$ (confidence calibration) against the widely used $\mathrm{ConfCE}_{\mathsf B}$ and $\mathrm{ConfECE}_{\mathsf B}$.

\subsection{Results}

% \begin{description}
%     \item[Dataset and Training] We use dataset \texttt{CIFAR-100} for evaluation. 
%     \item [Models] MobileNetV3-Small, ResNet10/18/34/50. We keep each checkpoint in the training process. We evaluate the model after temperature scaling as in \citet{minderer2021revisiting}. For temperature scaling, we select the temperature of the model output to minimize the log loss (a.k.a., cross-entropy loss) on the validation set. 
% \item [Evaluation Metrics] We report the log loss, the quadratic loss, the accuracy, the spherical loss, and calibration errors in \Cref{sec: truthful calibration measure}. 
% \end{description}
    \minspace{-2mm}
\paragraph{Assumption validation.}
Our theory applies to calibrated or nearly calibrated predictors. After temperature scaling, the evaluated checkpoints lie close to the calibrated-predictor loss lines for both the confidence and classwise measures, indicating that the models are approximately calibrated in the regime relevant to the theory. We defer these assumption-validation plots to \Cref{apdx: approx calibrate empirical}.

\begin{figure}[ht]
    \centering
    \begin{subfigure}[t]{0.49\linewidth}
        \centering
        \includegraphics[width=\linewidth]{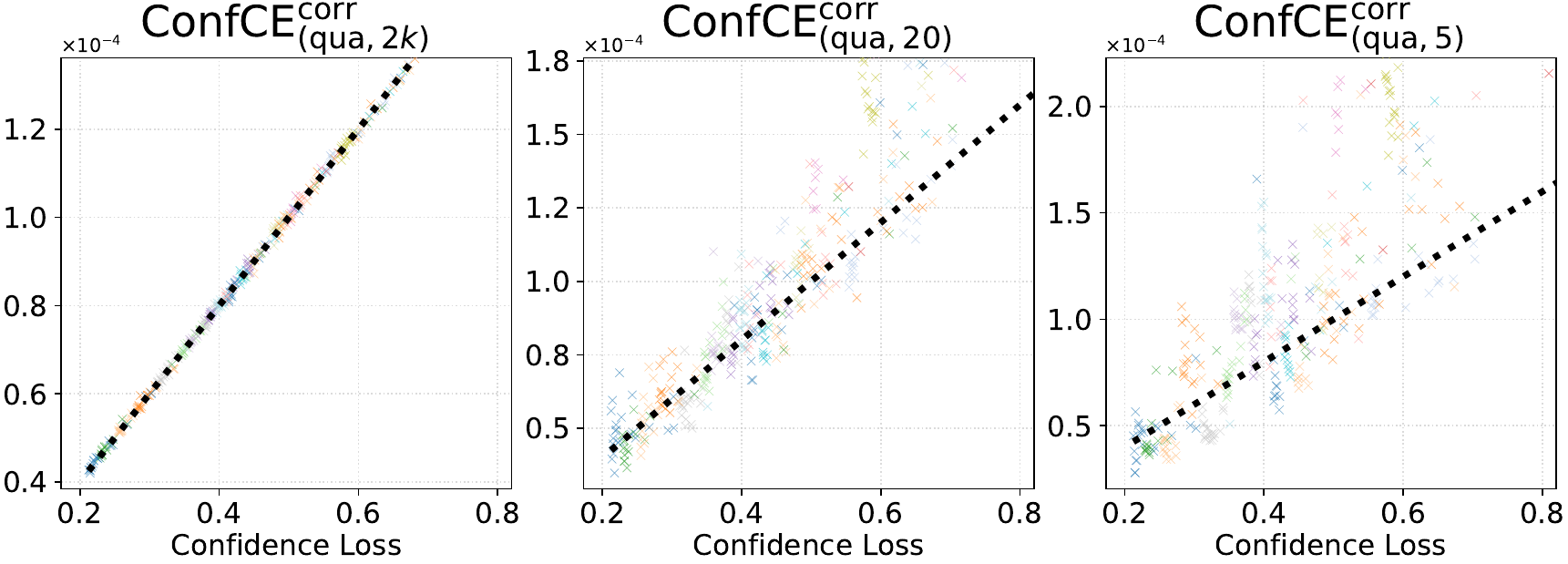}
        \caption{$\mathrm{ConfCE}^{\mathrm{corr}}_{\mathsf B}$ on the maximum dominance set.}
        \label{fig:dominance-a}
    \end{subfigure}
    \hfill
    \begin{subfigure}[t]{0.49\linewidth}
        \centering
        \includegraphics[width=\linewidth]{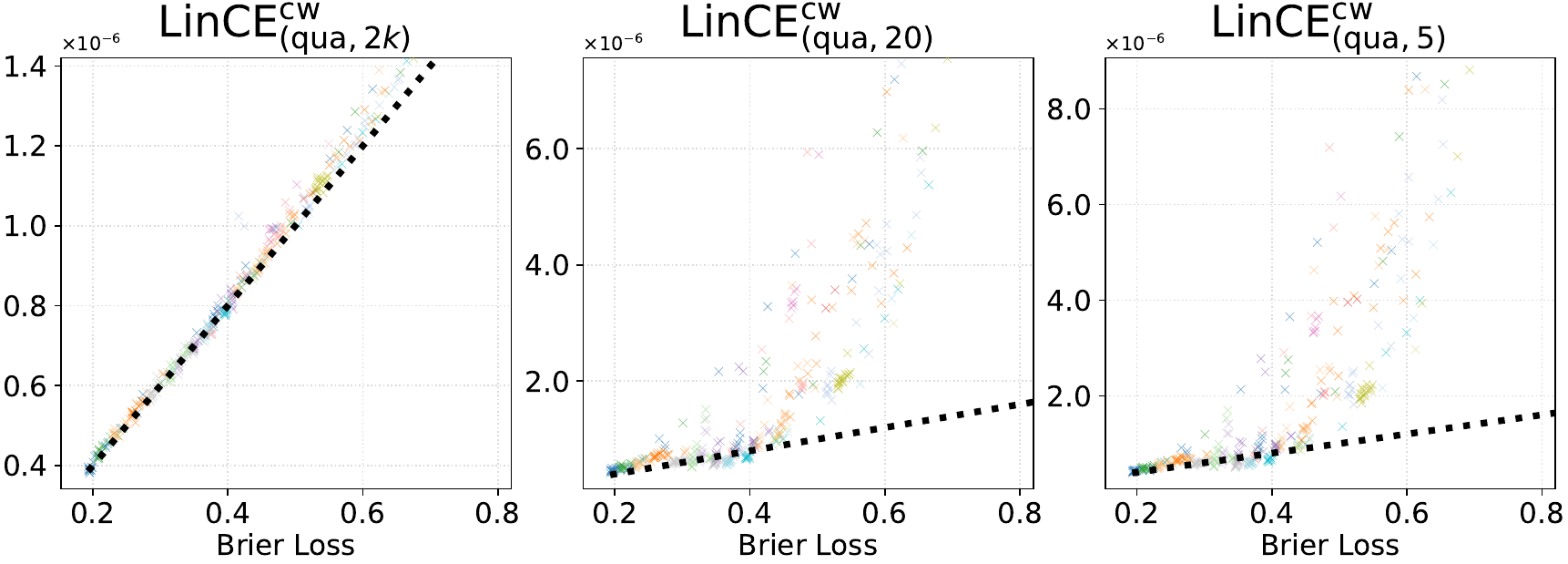}
        \caption{$\mathrm{LinCE}^{\mathrm{cw}}_{\mathsf B}$ on the maximum dominance set.}
        \label{fig:dominance-b}
    \end{subfigure}
     \minspace{-1mm}
    \caption{Empirical validation of dominance preservation on the cross-model maximum dominance set. Each panel visualizes only the predictors belonging to the extracted dominance total order; predictors outside this order are omitted. The plotted annotations show both the calibration errors of the retained predictors and the rank correlation between the dominance order and the order induced by the calibration error. Our truthful calibration errors achieve high rank correlation, and the few remaining inversions can be attributed to sampling error and the fact that the predictors are only nearly calibrated. The analogous within-model plots are deferred to the appendix.}
    \label{fig:dominance-ab}
 \minspace{-3mm}
\end{figure}

    \minspace{-2mm}
\paragraph{Dominance preservation.}
\Cref{thm:blackwell-general} predicts that a truthful calibration error should preserve the Blackwell ordering induced by proper losses. To visualize this, we compare all evaluated predictors under the family of proper losses described above, form the induced dominance partial order, and plot only its maximum dominance set, namely the largest subset that admits a total order. %\footnote{Equivalently, we extract a longest path in the dominance partial order.} 
\Cref{fig:dominance-ab} shows that both truthful calibration errors track this ordering closely. The ordering is stable across binning choices, and the small number of remaining inversions is consistent with sampling noise and slight residual miscalibration. The analogous within-model plots are deferred to \Cref{apdx: dominance preservation}.

\begin{figure}[htbp]
    \centering
    \begin{subfigure}[t]{0.49\linewidth}
        \centering
        \includegraphics[width=\linewidth]{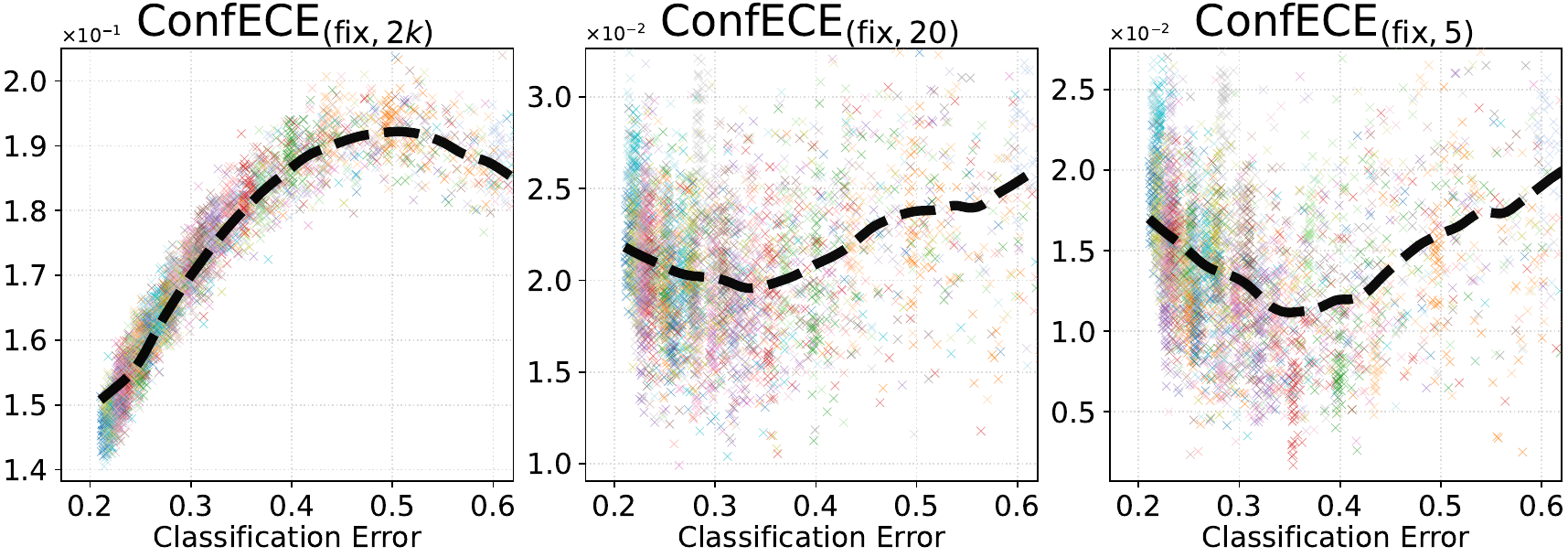}
        \caption{Non-truthful $\mathrm{ConfECE}_{\mathsf B}$.}
        \label{fig:trend-a}
    \end{subfigure}
    \hfill
    \begin{subfigure}[t]{0.49\linewidth}
        \centering
        \includegraphics[width=\linewidth]{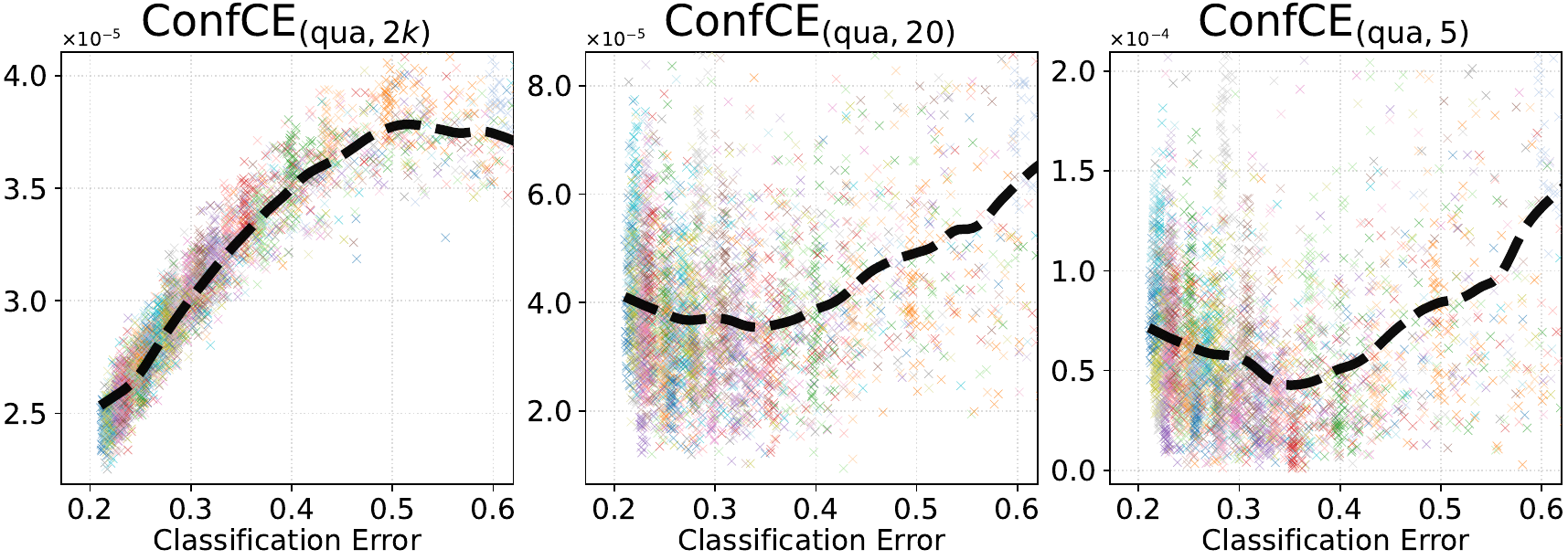}
        \caption{Non-truthful $\mathrm{ConfCE}_{\mathsf B}$.}
        \label{fig:trend-b}
    \end{subfigure}
\vspace{0.8em}
    \begin{subfigure}[t]{0.49\linewidth}
        \centering
        \includegraphics[width=\linewidth]{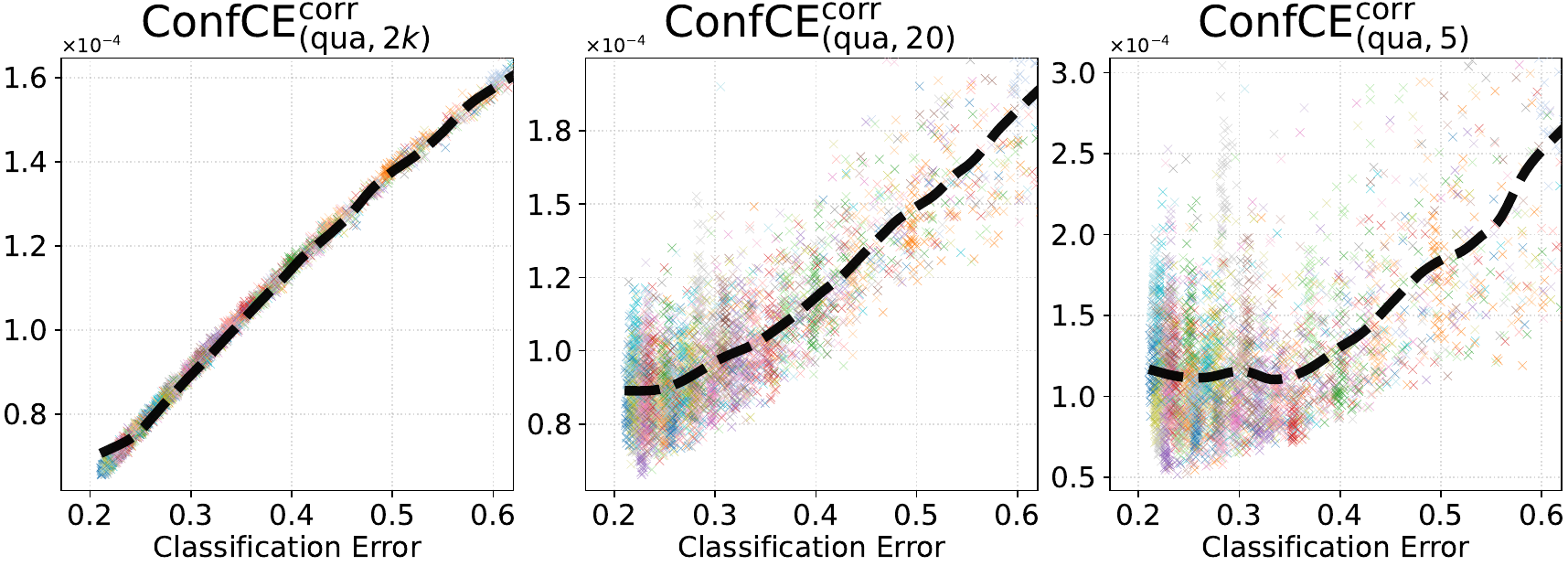}
        \caption{Truthful $\mathrm{ConfCE}^{\mathrm{corr}}_{\mathsf B}$.}
        \label{fig:trend-c}
    \end{subfigure}
    \hfill
    \begin{subfigure}[t]{0.49\linewidth}
        \centering
        \includegraphics[width=\linewidth]{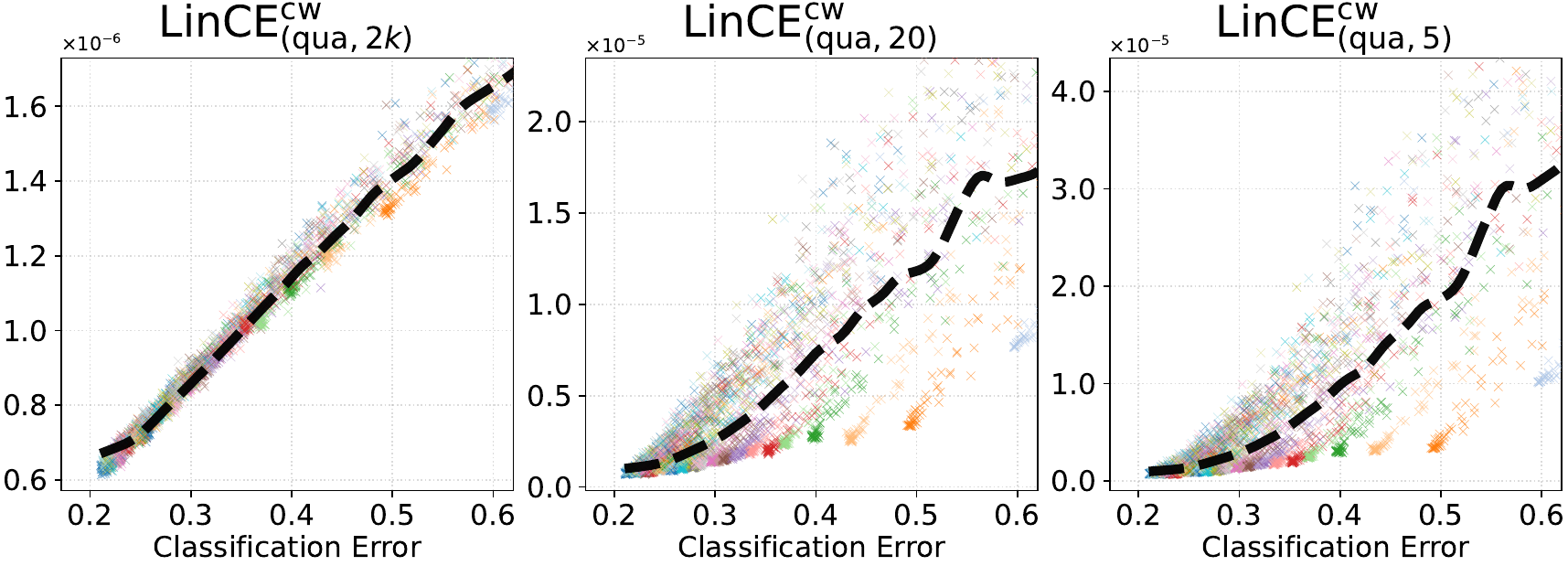}
        \caption{Truthful $\mathrm{LinCE}^{\mathrm{cw}}$.}
        \label{fig:trend-d}
    \end{subfigure}
 \minspace{-1mm}
    \caption{Binning Robustness on the within-model MobileNetV3-Small-1.0 traces. Each panel plots calibration error against classification error for temperature-scaled checkpoints, and the curves are kernel-regression trend lines for different bin sizes. The truthful calibration errors in (c) and (d) remain monotone: lower-loss predictors also have lower calibration error across binning choices. By contrast, the non-truthful calibration errors in (a) and (b) can flip their relationship with predictive loss as the binning size changes. }
     \minspace{-2mm}
    \label{fig:trend-abcd}
    % \minspace{-2mm}
\end{figure}

    \minspace{-2mm}
\paragraph{Binning robustness.}
\label{sec: empirical per model}
Prior empirical work reported that the relationship between calibration and predictive performance can flip when only the number of bins changes. We now move beyond ranking preservation on the Blackwell-dominated subset and quantitatively examine all predictors in the within-model traces. \Cref{fig:trend-abcd} shows that this instability is concentrated in the non-truthful confidence-based errors: their curves can rise again late in training, making better predictors appear less calibrated. By contrast, for the truthful calibration errors $\mathrm{ConfCE}^{\mathrm{corr}}_{\mathsf B}$ and $\mathrm{LinCE}^{\mathrm{cw}}_{\mathsf B}$, the trend is stable across binning choices, with lower classification error continuing to align with lower calibration error. Thus, truthful errors are quantitatively more robust to the choice of binning size. This supports our interpretation that the observed flip is a measurement artifact of non-truthful calibration errors rather than a property of the underlying predictors. We provide the analogous cross-model plots in the appendix, and \Cref{tab:trend-spearman} reports the corresponding rank correlations across binning choices.

We discuss the limitations in \Cref{sec: limitations}.

%%%%%%%%%%%%%%%%%%%%%%%%%%%%%%%%%%%%%%%%%%%%%%%%%%%%%%%%%%%%

\appendix

\section{Limitations}
\label{sec: limitations}

Our theoretical guarantees apply under the assumptions stated in the analysis,  that the predictors are nearly calibrated for the relevant notion of calibration. In particular, our approximate ranking result assumes that the remaining calibration bias is small relative to the gap in the associated downstream loss. Thus, the theory does not claim that truthful calibration errors recover decision-theoretic rankings for arbitrary predictors; it characterizes the regime in which calibration errors can be interpreted as stable proxies for downstream decision quality.

Our experiments are designed to validate that this regime is realistic under the standard post-hoc calibration pipeline in prior work. %Following prior work on neural-network calibration \citep{minderer2021revisiting}, we apply temperature scaling to classifier checkpoints before evaluation, and empirically observe that the resulting predictors lie close to the calibrated-predictor loss lines. This supports the use of the nearly calibrated assumption in our experimental setting and is consistent with the standard practice of first post-hoc calibrating probabilistic predictors before comparing calibration errors.
The results therefore should not be interpreted as applying unchanged to settings where the predictor is substantially miscalibrated. This limitation is especially relevant for post-trained language models, whose reported probabilities or confidence scores can remain poorly calibrated after instruction tuning, reinforcement learning, or other post-training procedures, e.g., \citep{achiam2023gpt}. In such regimes, the calibration-bias term may dominate the downstream loss gap, and the ranking guarantees in this paper need not hold. 
%Extending truthful calibration measurement to substantially miscalibrated predictors, or combining our errors with stronger recalibration methods for post-trained language models, is an important direction for future work.

% Our empirical study is also intentionally narrow in scope. We focus on
% post-hoc temperature-scaled image classifiers on CIFAR-100, and the main text
% emphasizes quantile binning because this is the setting covered by the
% theoretical truthfulness results. Although the observed trends are consistent
% across the within-model and cross-model comparisons we report, broader evidence
% across more datasets, calibration methods, and deployment settings would be
% needed before treating the robustness patterns here as universal. Extending the
% analysis to richer calibration notions and to estimators beyond the current
% binning-based framework remains an important direction for future work.

\section{Additional Preliminaries}
\subsection{Calibration Measure: Completeness and Soundness}
\label{apdx: c and s}

% \lunjia{todo}
% A calibration measure over $\nsp$ samples is a mapping $\CAL: [\Delta(\truthsp)]^\nsp\times\truthsp^\nsp\to\reals$ from $\nsp$ predictions and $\nsp$ outcomes to a real-valued error. A commonly used calibration measure is the \emph{Expected Calibration Error (ECE)}:

% A calibration error is truthful if the expected error is minimized when a predictor outputs the true underlying distribution of outcomes. 

% \begin{definition}[Truthful]
% \label{def: truthful}
%     A calibration measure for $\nsp$ samples is \emph{truthful} if for 
%     \begin{equation*}
%         \expect{}{\CAL(\left(f(x_i)\right)_{i=1}^\nsp, \left(y_i\right)_{i=1}^\nsp)} 
%         \;\;\text{is minimized when } f(x) = \Pr[\truth \mid \feature = x].
%     \end{equation*}
% \end{definition}

The two requirements for a calibration measure are
\emph{completeness} and \emph{soundness}. Completeness and soundness ensure
that the calibration error vanishes only for calibrated predictors and does
not vanish for miscalibrated predictors. Note that proper losses are truthful,
but are not calibration errors, i.e., they do not have completeness or
soundness as calibration measures.
% These properties concern the behavior of the calibration error as the sample size grows. 

\begin{definition}[Completeness]
    A sequence of calibration measures $\{\CAL_n\}_{n\ge1}$ is
    \emph{complete in the limit} if, for any calibrated predictor $f$,
    \begin{equation*}
        \lim_{\nsp \to \infty} 
        \; \expect{}{\CAL_n\!\left(\left(f(x_i)\right)_{i=1}^\nsp, \left(y_i\right)_{i=1}^\nsp\right)} 
        = 0.
    \end{equation*}
\end{definition}

\begin{definition}[Soundness]
    A sequence of calibration measures $\{\CAL_n\}_{n\ge1}$ is
    \emph{sound in the limit} if, for any miscalibrated predictor $f$,
    \begin{equation*}
        \liminf_{\nsp \to \infty}
        \; \expect{}{\CAL_n\!\left(\left(f(x_i)\right)_{i=1}^\nsp, \left(y_i\right)_{i=1}^\nsp\right)} 
        > 0.
    \end{equation*}
\end{definition}

In this paper, we focus on several weaker definitions of high-dimensional
calibration. The classwise aggregation focuses on classwise calibration
\citep{class-wise}. Instead of conditioning on the high-dimensional
prediction, classwise calibration requires unbiasedness conditioned on a
prediction of a single dimension.

\begin{definition}[Classwise Calibration \citep{class-wise}]

\label{def: cal classwise}
    A predictor $f:X\to \Delta(\truthsp)$ is (perfectly)  classwise \emph{calibrated} if for every $p\in [0, 1]$ and every $r \in \truthsp = \{1,\ldots,\ndim\}$,
    \[\Pr_{(x,y)\sim D}[\truth = r | \classifier_r(\feature) = \pred] = \pred.\]

\end{definition}

The confidence aggregation is induced from confidence calibration. Confidence
calibration requires unbiasedness conditioned on the highest prediction.

\begin{definition}[Confidence Calibration]

\label{def: cal confidence}
    A predictor $f:X\to \Delta(\truthsp)$ is (perfectly) confidence \emph{calibrated} if for every $p\in [0, 1]$,
    \[\Pr_{(x,y)\sim D}[\truth = \argmax_r \classifier_r(\feature) | \max_r \classifier_r(\feature) = \pred] = \pred,\]
    where $\argmax$ uses the fixed deterministic tie-breaking rule from the main text.

\end{definition}

\subsection{Commonly Used Proper Losses}
\label{apdx: proper loss example}
We define the Brier loss and the classification error here. 
\begin{definition}[Brier Loss]
    The multiclass Brier loss is
    \begin{equation*}
        l_{\mathrm{Brier}}(p,y):=  \sum_{r=1}^k \bigl(p\sps r -\ind{y = r}\bigr)^2 .
    \end{equation*}
\end{definition}

\begin{definition}
    [Classification Error] The multiclass classification error is
    \begin{equation*}
        l_{\mathrm{Cla}}(p,y):= \ind{y \ne \argmax_{r\in \{1,\ldots,\ndim\}}p_r}.
    \end{equation*}
\end{definition}

\section{Proofs for \Cref{sec:truthful}}
\label{apdx:truthful}

\begin{proof}[Proof of \Cref{thm:linear-property-truthful}]
Fix arbitrary reports $u_{1:n}\in\mathcal R_\phi^n$, and let
\[
    \mu_i^* := \Gamma_\phi(P_i)=\E_{y_i\sim P_i}[\phi(y_i)]
\]
for $i=1,\ldots,n$.  Because the binning rule $\mathsf B$ is
outcome-independent, conditioning on $u_{1:n}$ fixes the partition
$\mathsf B(u_{1:n})$.  For each $i$, define the centered noise
\[
    \xi_i := \phi(y_i)-\mu_i^*.
\]
Then $\E[\xi_i]=0$, and the variables $\xi_1,\ldots,\xi_n$ are
independent.

For any bin $B\in\mathsf B(u_{1:n})$,
\[
    \sum_{i\in B}(u_i-\phi(y_i))
    =
    \sum_{i\in B}(u_i-\mu_i^*)
    -
    \sum_{i\in B}\xi_i.
\]
Taking squared Euclidean norms and expectations, the cross term vanishes
because the second sum has mean zero.  Independence also removes the
cross-covariances among the $\xi_i$, so
\[
\begin{aligned}
    \E\!\left[
        \left\|
            \sum_{i\in B}(u_i-\phi(y_i))
        \right\|_2^2
    \right]
    &=
    \left\|
        \sum_{i\in B}(u_i-\mu_i^*)
    \right\|_2^2
    +
    \sum_{i\in B}
    \E\big[\|\xi_i\|_2^2\big] \\
    &=
    \left\|
        \sum_{i\in B}(u_i-\mu_i^*)
    \right\|_2^2
    +
    \sum_{i\in B}
    \E_{y_i\sim P_i}\big[\|\phi(y_i)-\mu_i^*\|_2^2\big].
\end{aligned}
\]
Summing over bins and dividing by $n^2$ yields
\[
\begin{aligned}
    \E\big[
    \mathrm{LinCE}_{\phi,\mathsf B}
    (u_{1:n};y_{1:n})
    \big]
    & =
    \frac{1}{n^2}
    \sum_{B\in\mathsf B(u_{1:n})}
    \left\|
        \sum_{i\in B}(u_i-\mu_i^*)
    \right\|_2^2 \\
    &\qquad
    +
    \frac{1}{n^2}
    \sum_{i=1}^n
    \E_{y_i\sim P_i}\big[\|\phi(y_i)-\mu_i^*\|_2^2\big].
\end{aligned}
\]
The second term is independent of the report, while the first term is
nonnegative and vanishes at the truthful report $u_i=\mu_i^*$ for all
$i$.  This proves truthfulness.

For the final scalar claim, suppose $d=1$ and let
$\mathcal R_\phi=[a,b]$.  If $a=b$, then $\Gamma_\phi$ is constant, so
calibration is trivial and the error is identically zero.  Otherwise,
define the affine rescaling
\[
    \tilde\phi(y):=\frac{\phi(y)-a}{b-a}
    \qquad\text{and}\qquad
    \tilde u_i:=\frac{u_i-a}{b-a}.
\]
Then $\tilde\phi$ takes values in $[0,1]$, quantile bins are unchanged
by this monotone rescaling, and
\[
    \mathrm{LinCE}_{\phi,\mathsf B_n}(u_{1:n};y_{1:n})
    =
    (b-a)^2\,
    \mathrm{LinCE}_{\tilde\phi,\mathsf B_n}(\tilde u_{1:n};y_{1:n}).
\]
Thus $\{\mathrm{LinCE}_{\phi,\mathsf B_n}\}_{n\ge1}$ differs from the
scalar squared quantile-binned calibration error of
\citet{hartline2025perfectly} only by a deterministic positive factor.
Under the regime $\mathsf B_n=\mathsf B_{m_n}^Q$ with $m_n\to\infty$ and
$n/m_n\to\infty$, their completeness and soundness result applies to the
rescaled problem, and positive rescaling preserves those limit
properties.  Hence
$\{\mathrm{LinCE}_{\phi,\mathsf B_n}\}_{n\ge1}$ is complete and sound in
the limit for calibration of $\Gamma_\phi$.
\end{proof}

\begin{example}[Confidence aggregation can be non-truthful]
\label{apdx:confidence-nontruthful-example}
Let $k\ge 3$, and let $j$ be the label selected by the fixed deterministic
tie-breaking rule when the reported distribution is uniform.  Choose two
distinct labels $s,t\neq j$, and suppose the true posterior is the same on
every sample:
\[
    p_i^* = p^*
    :=
    \frac{1}{2}e_s + \frac{1}{k}e_j + \left(\frac{1}{2}-\frac{1}{k}\right)e_t .
\]
Thus class $s$ is uniquely most likely, while class $j$ has true probability
$1/k$.  Consider applying the truthful binary calibration error
$\mathrm{LinCE}_{\mathrm{id},\mathsf B}$ to the confidence pairs
$(c_i,z_i)$ produced by confidence aggregation.

Under truthful reporting, each selected confidence equals $c_i=1/2$, and the
selected class is correct with probability $1/2$, so
$z_i\sim\mathrm{Bernoulli}(1/2)$.  Since all reports fall in the same bin, the
expected error is
\[
    \E\!\left[
        \mathrm{LinCE}_{\mathrm{id},\mathsf B}(c_{1:n};z_{1:n})
    \right]
    =
    \frac{1}{n^2}\Var\!\left[\sum_{i=1}^n z_i\right]
    =
    \frac{1}{4n}.
\]

If instead the predictor always reports the uniform distribution
$(1/k,\ldots,1/k)$, deterministic tie-breaking selects class $j$.  Then
$c_i=1/k$ and $z_i\sim\mathrm{Bernoulli}(1/k)$, again with all reports in one
bin, so
\[
    \E\!\left[
        \mathrm{LinCE}_{\mathrm{id},\mathsf B}(c_{1:n};z_{1:n})
    \right]
    =
    \frac{1}{n^2}\Var\!\left[\sum_{i=1}^n z_i\right]
    =
    \frac{1}{n}\frac{1}{k}\left(1-\frac{1}{k}\right)
    <
    \frac{1}{4n}.
\]
Thus confidence aggregation can assign smaller expected error to a
non-truthful report, even when the underlying binary calibration error
is truthful.
\end{example}

\begin{proof}[Proof of \Cref{thm:confidence-truthful}]
Let $p_1^*,\ldots,p_n^*\in\Delta(\truthsp)$ be the true class
distributions and draw $y_i\sim p_i^*$ independently.  Fix arbitrary
reported distributions $p_1,\ldots,p_n$, and write
\[
    r_i:=\argmax_{r\in\truthsp} p_i^{(r)},
    \qquad
    c_i:=p_i^{(r_i)},
    \qquad
    q_i:=(p_i^*)^{(r_i)},
    \qquad
    z_i:=\ind{y_i=r_i}.
\]
Conditional on the reports, the selected classes $r_i$, confidences
$c_i$, and bins $\mathsf B(c_{1:n})$ are fixed, while the labels
$z_i\sim\mathrm{Bernoulli}(q_i)$ remain independent.

For any bin $B\in\mathsf B(c_{1:n})$,
\[
    \sum_{i\in B}(c_i-z_i)
    =
    \sum_{i\in B}(c_i-q_i)
    -
    \sum_{i\in B}(z_i-q_i).
\]
The second sum has conditional mean zero, so as in the previous proof,
\[
\begin{aligned}
    \E\!\left[
        \left(
            \sum_{i\in B}(c_i-z_i)
        \right)^2
        \middle|\, p_{1:n}
    \right]
    &=
    \left(
        \sum_{i\in B}(c_i-q_i)
    \right)^2
    +
    \sum_{i\in B} q_i(1-q_i).
\end{aligned}
\]
Summing over bins gives
\[
\begin{aligned}
    &\E\!\left[
        \frac{1}{n^2}
        \sum_{B\in\mathsf B(c_{1:n})}
        \left(
            \sum_{i\in B}(c_i-z_i)
        \right)^2
        \middle|\, p_{1:n}
    \right] \\
    &\qquad=
    \frac{1}{n^2}
    \sum_{B\in\mathsf B(c_{1:n})}
    \left(
        \sum_{i\in B}(c_i-q_i)
    \right)^2
    +
    \frac{1}{n^2}
    \sum_{i=1}^n q_i(1-q_i).
\end{aligned}
\]
The correction term contributes
\[
\begin{aligned}
    \E\!\left[
        \frac{1}{n}
        \left(
            1-\frac{1}{n}\sum_{i=1}^n z_i
        \right)
        \middle|\, p_{1:n}
    \right]
    =
    \frac{1}{n^2}\sum_{i=1}^n (1-q_i).
\end{aligned}
\]
Therefore
\[
\begin{aligned}
    n^2\,
    \E\big[
        \mathrm{ConfCE}^{\mathrm{corr}}_{\mathsf B}(p_{1:n};y_{1:n})
        \mid p_{1:n}
    \big]
    & =
    \sum_{B\in\mathsf B(c_{1:n})}
    \left(
        \sum_{i\in B}(c_i-q_i)
    \right)^2
    +
    \sum_{i=1}^n (1-q_i^2).
\end{aligned}
\]
The first term is nonnegative and, for fixed selected classes
$r_1,\ldots,r_n$, is minimized by setting $c_i=q_i$ for every $i$.
The second term depends only on the selected classes and is minimized by
maximizing each $q_i$, that is, by choosing
$r_i\in\argmax_{r\in\truthsp}(p_i^*)^{(r)}$.  The truthful report
$p_i=p_i^*$ attains both conditions simultaneously (with deterministic
tie-breaking selecting one maximizer when several classes tie).  Hence no
alternative report has smaller expected error, proving truthfulness.

For limiting completeness and soundness, fix a predictor
$h:\featuresp\to\Delta(\truthsp)$ and let
\[
    c(x):=\max_{r\in\truthsp} h^{(r)}(x),
    \qquad
    z:=\ind{y=r_h(x)},
\]
where $r_h(x)$ is the top label selected by the fixed deterministic
tie-breaking rule.  Along a consistent quantile regime
$\mathsf B_n=\mathsf B_{m_n}^Q$, the corrected confidence error decomposes as
\[
    \mathrm{ConfCE}^{\mathrm{corr}}_{\mathsf B_n}(h(x_{1:n});y_{1:n})
    =
    \mathrm{LinCE}_{\mathrm{id},\mathsf B_n}(c(x_{1:n});z_{1:n})
    +
    \frac{1}{n}
    \left[
        1-\frac{1}{n}\sum_{i=1}^n z_i
    \right].
\]
The correction term is always between $0$ and $1/n$, so it vanishes in
expectation as $n\to\infty$.  The remaining term is exactly the scalar squared
quantile-binned calibration error for the induced binary prediction problem
$(c,z)$.  Moreover, $h$ is confidence calibrated if and only if the scalar
predictor $c$ is calibrated for the binary outcome $z$.  Therefore Hartline
et al.'s scalar completeness and soundness result applies to
$\mathrm{LinCE}_{\mathrm{id},\mathsf B_n}(c(x_{1:n});z_{1:n})$, and adding the
vanishing correction preserves the same limit properties.  Hence
$\{\mathrm{ConfCE}^{\mathrm{corr}}_{\mathsf B_n}\}_{n\ge1}$ is complete and
sound in the limit for confidence calibration.
\end{proof}

\section{Proofs for \Cref{sec:blackwell}}
\label{apdx:general-blackwell}

\subsection{Proof of \Cref{thm:blackwell-general}}
\begin{proof}[Proof of \Cref{thm:blackwell-general}]
Let
\[
    F_i := f(x_i)
    \qquad\text{and}\qquad
    G_i := g(x_i)
\]
for $i=1,\ldots,n$. Since $f\succeq_B g$, there is a Markov kernel from
$F_i$ to $G_i$, so $G_i$ is generated from $F_i$ without observing $y_i$.
Because $(x_i,y_i)_{i=1}^n$ are drawn i.i.d., conditional on
$F_{1:n},G_{1:n}$ the outcomes $y_1,\ldots,y_n$ remain independent.

Since $f$ is calibrated for $\Gamma_\phi$,
\[
    \E[\phi(y_i)\mid F_i] = F_i.
\]
Because $G_i$ is sampled from $F_i$ without access to $y_i$, conditioning on
$G_i$ does not change this conditional property value:
\[
    \E[\phi(y_i)\mid F_{1:n},G_{1:n}] = \E[\phi(y_i)\mid F_i] = F_i.
\]
Thus, after conditioning on $F_{1:n},G_{1:n}$, the $n$ conditional outcome
distributions have truthful reports $F_1,\ldots,F_n$, while
$G_1,\ldots,G_n$ are alternative reports in $\mathcal R_\phi$. By the
truthfulness of $\CAL_\phi$,
\[
    \E\!\left[
        \CAL_\phi(F_{1:n};y_{1:n})
        \,\middle|\,
        F_{1:n},G_{1:n}
    \right]
    \le
    \E\!\left[
        \CAL_\phi(G_{1:n};y_{1:n})
        \,\middle|\,
        F_{1:n},G_{1:n}
    \right].
\]
Taking expectation over $F_{1:n},G_{1:n}$ proves the claim.
\end{proof}

\subsection{Proof of \Cref{thm:recalibration-general}}
\begin{proof}[Proof of \Cref{thm:recalibration-general}]
Let $U_i := h(x_i)$ and $\bar U_i := \bar h(x_i)$ for $i=1,\ldots,n$, where
\[
    \bar U_i = \E[\phi(y_i)\mid U_i].
\]
Because the samples are i.i.d., conditional on $U_{1:n}$ the outcomes
$y_1,\ldots,y_n$ are independent, and the truthful report for the $i$-th
conditional distribution is exactly $\bar U_i$. Therefore, by truthfulness of
$\CAL_\phi$,
\[
    \E\!\left[
        \CAL_\phi(\bar U_{1:n};y_{1:n})
        \,\middle|\,
        U_{1:n}
    \right]
    \le
    \E\!\left[
        \CAL_\phi(U_{1:n};y_{1:n})
        \,\middle|\,
        U_{1:n}
    \right].
\]
Taking expectation over $U_{1:n}$ yields
\[
    \E\!\left[
        \CAL_\phi(\bar h(x_{1:n});y_{1:n})
    \right]
    \le
    \E\!\left[
        \CAL_\phi(h(x_{1:n});y_{1:n})
    \right].
\qedhere
\]
\end{proof}

\subsection{Proof of Associated Losses}
\label{apdx:proper-loss-identities}

\begin{lemma}[Proper-loss identities for the constructed errors]
\label{lem:constructed-error-proper-loss}
Let $(x_i,y_i)_{i=1}^n$ be drawn i.i.d.\ from $D$.
\begin{enumerate}[leftmargin=*]
    \item If $h:\featuresp\to\mathcal R_\phi$ is calibrated for $\Gamma_\phi$, then for every outcome-independent binning rule $\mathsf B$,
    \[
        \E\!\left[
            \mathrm{LinCE}_{\phi,\mathsf B}
            (h(x_{1:n});y_{1:n})
        \right]
        =
        \frac{1}{n}
        \E_{(x,y)\sim D}\big[\ell_\phi(h(x),y)\big].
    \]
    \item If $h:\featuresp\to\Delta(\truthsp)$ is classwise calibrated, then for every outcome-independent binning rule $\mathsf B$,
    \[
        \E\!\left[
            \mathrm{LinCE}^{\mathrm{cw}}_{\mathsf B}
            (h(x_{1:n});y_{1:n})
        \right]
        =
        \frac{1}{n}
        \E_{(x,y)\sim D}\big[\ell_{\mathrm{Brier}}(h(x),y)/k\big].
    \]
    \item If $h:\featuresp\to\Delta(\truthsp)$ is confidence calibrated, then for every outcome-independent binning rule $\mathsf B$,
    \[
        \E\!\left[
            \mathrm{ConfCE}^{\mathrm{corr}}_{\mathsf B}
            (h(x_{1:n});y_{1:n})
        \right]
        =
        \frac{1}{n}
        \E_{(x,y)\sim D}\big[\ell_{\mathrm{conf}}(h(x),y)\big].
    \]
\end{enumerate}
\end{lemma}

\begin{proof}
We prove the three identities in order.

For the linear-property identity, let $U_i:=h(x_i)$.  Since $h$ is
calibrated for $\Gamma_\phi$,
\[
    \E[\phi(y_i)\mid U_i]=U_i
    \qquad\text{for each }i.
\]
Conditioning on $U_{1:n}$ fixes the bins of $\mathsf B$, because the
binning rule is outcome-independent.  Therefore the centered residuals
$U_i-\phi(y_i)$ have conditional mean zero and are independent across
$i$, so the cross terms vanish:
\[
\begin{aligned}
    \E\!\left[
        \mathrm{LinCE}_{\phi,\mathsf B}(h(x_{1:n});y_{1:n})
        \mid U_{1:n}
    \right]
    &=
    \frac{1}{n^2}
    \sum_{i=1}^n
    \E\!\left[
        \|U_i-\phi(y_i)\|_2^2
        \mid U_i
    \right].
\end{aligned}
\]
Taking expectation and using that the samples are i.i.d. gives
\[
\begin{aligned}
    \E\!\left[
        \mathrm{LinCE}_{\phi,\mathsf B}(h(x_{1:n});y_{1:n})
    \right]
    &=
    \frac{1}{n}
    \E_{(x,y)\sim D}\big[\|h(x)-\phi(y)\|_2^2\big] \\
    &=
    \frac{1}{n}
    \E_{(x,y)\sim D}\big[\ell_\phi(h(x),y)\big].
\end{aligned}
\]

For the classwise identity, fix a class $r\in\{1,\ldots,k\}$ and define
$U_i^{(r)}:=h^{(r)}(x_i)$ and $Z_i^{(r)}:=\ind{y_i=r}$.  Classwise
calibration gives
\[
    \E[Z_i^{(r)}\mid U_i^{(r)}]=U_i^{(r)}.
\]
Applying the linear-property identity above to the scalar property
$\phi_r(y)=\ind{y=r}$ and the binning rule $\mathsf B$ yields
\[
    \E\!\left[
        \mathrm{LinCE}_{\phi_r,\mathsf B}
        (h^{(r)}(x_{1:n});y_{1:n})
    \right]
    =
    \frac{1}{n}
    \E_{(x,y)\sim D}\big[(h^{(r)}(x)-\ind{y=r})^2\big].
\]
Averaging over $r=1,\ldots,k$ and using the definition of
$\mathrm{LinCE}^{\mathrm{cw}}_{\mathsf B}$ gives
\[
\begin{aligned}
    \E\!\left[
        \mathrm{LinCE}^{\mathrm{cw}}_{\mathsf B}(h(x_{1:n});y_{1:n})
    \right]
    &=
    \frac{1}{kn}
    \sum_{r=1}^k
    \E_{(x,y)\sim D}\big[(h^{(r)}(x)-\ind{y=r})^2\big] \\
    &=
    \frac{1}{n}
    \E_{(x,y)\sim D}\big[\ell_{\mathrm{Brier}}(h(x),y)/k\big].
\end{aligned}
\]

For the confidence identity, let
\[
    r_i:=\argmax_{r\in\truthsp} h^{(r)}(x_i),
    \qquad
    c_i:=h^{(r_i)}(x_i),
    \qquad
    z_i:=\ind{y_i=r_i}.
\]
Confidence calibration implies
\[
    \E[z_i\mid c_i]=c_i.
\]
Conditioning on $c_{1:n}$ fixes the bins of $\mathsf B(c_{1:n})$, and the
variables $z_1,\ldots,z_n$ remain independent.  Hence
\[
\begin{aligned}
    &\E\!\left[
        \frac{1}{n^2}
        \sum_{B\in\mathsf B(c_{1:n})}
        \left(
            \sum_{i\in B}(c_i-z_i)
        \right)^2
        \middle|\, c_{1:n}
    \right] \\
    &\qquad=
    \frac{1}{n^2}
    \sum_{i=1}^n c_i(1-c_i).
\end{aligned}
\]
The correction term contributes
\[
    \E\!\left[
        \frac{1}{n}
        \left(
            1-\frac{1}{n}\sum_{i=1}^n z_i
        \right)
        \middle|\, c_{1:n}
    \right]
    =
    \frac{1}{n^2}\sum_{i=1}^n (1-c_i).
\]
Adding the two pieces gives
\[
    \E\!\left[
        \mathrm{ConfCE}^{\mathrm{corr}}_{\mathsf B}(h(x_{1:n});y_{1:n})
        \middle|\, c_{1:n}
    \right]
    =
    \frac{1}{n^2}\sum_{i=1}^n (1-c_i^2).
\]
On the other hand,
\[
    \ell_{\mathrm{conf}}(h(x_i),y_i)
    =
    1-z_i+(c_i-z_i)^2,
\]
so by confidence calibration,
\[
    \E\big[\ell_{\mathrm{conf}}(h(x_i),y_i)\mid c_i\big]
    =
    1-c_i^2.
\]
Taking expectations and using i.i.d. sampling yields
\[
\begin{aligned}
    \E\!\left[
        \mathrm{ConfCE}^{\mathrm{corr}}_{\mathsf B}(h(x_{1:n});y_{1:n})
    \right]
    &=
    \frac{1}{n}
    \E_{(x,y)\sim D}\big[\ell_{\mathrm{conf}}(h(x),y)\big].
\end{aligned}
\qedhere
\]
\end{proof}

\subsection{Proof of \Cref{thm:approx-constructed-order}}
\label{apdx:approx-constructed}

\begin{proof}
Using the decomposition
\[
    n\,\E[\CAL(h)]
    =
    \E\!\left[
        \ell_\CAL(\bar h(x),y)
    \right]
    +
    \rho_\CAL(h),
\]
we obtain
\[
\begin{aligned}
    n\,\E[\CAL(f)]-n\,\E[\CAL(g)]
    &=
    \E[\ell_\CAL(\bar f(x),y)]
    -
    \E[\ell_\CAL(\bar g(x),y)]
    +
    \rho_\CAL(f)-\rho_\CAL(g) \\
    &\le
    -\epsilon + \epsilon
    \le 0,
\end{aligned}
\]
where the first inequality uses the assumed proper-loss separation together with
$\rho_\CAL(f)\le\epsilon$ and $\rho_\CAL(g)\ge 0$.  Dividing by $n$
proves the theorem.
\end{proof}

\section{Additional Empirical Results}
\label{apdx: additional empirical}

\subsection{Assumption Validation: Predictors are Approximately Calibrated}
\label{apdx: approx calibrate empirical}

\begin{figure}[ht]
    \centering
    \begin{subfigure}[t]{0.49\linewidth}
        \centering
        \includegraphics[width=\linewidth]{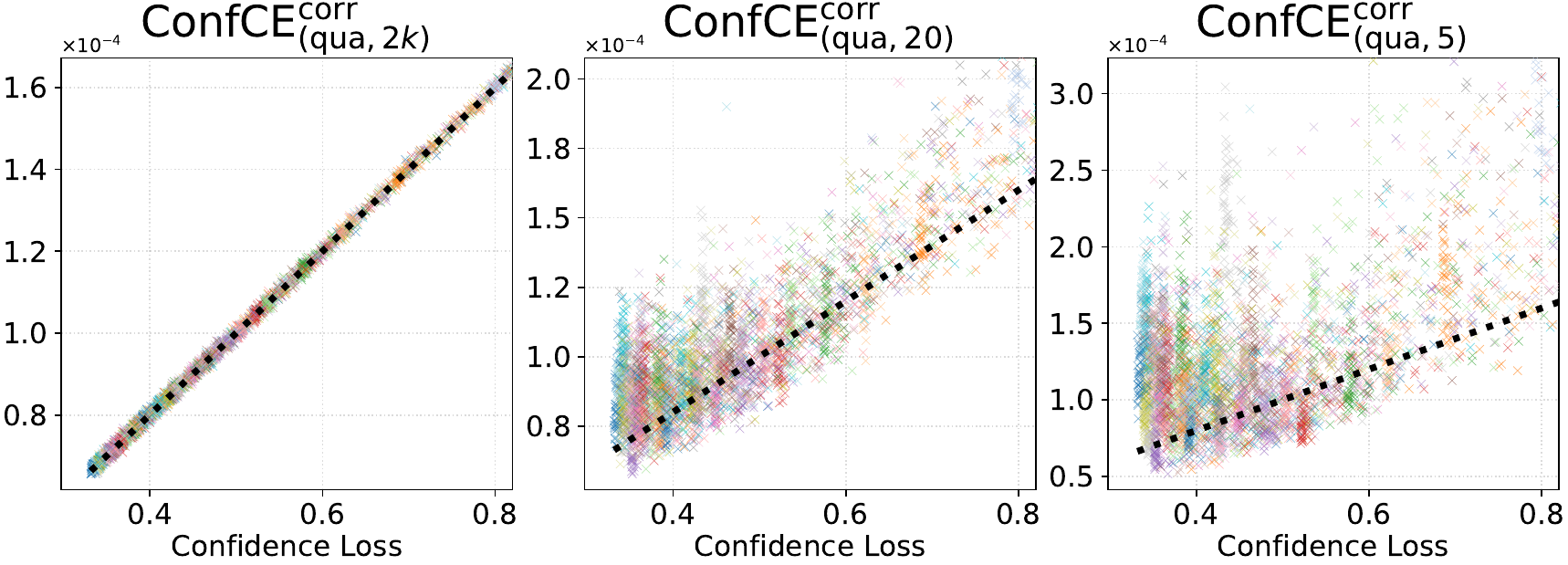}
        \caption{Within-model, confidence calibration.}
        \label{fig:estimate-loss-within-conf}
    \end{subfigure}
    \hfill
    \begin{subfigure}[t]{0.49\linewidth}
        \centering
        \includegraphics[width=\linewidth]{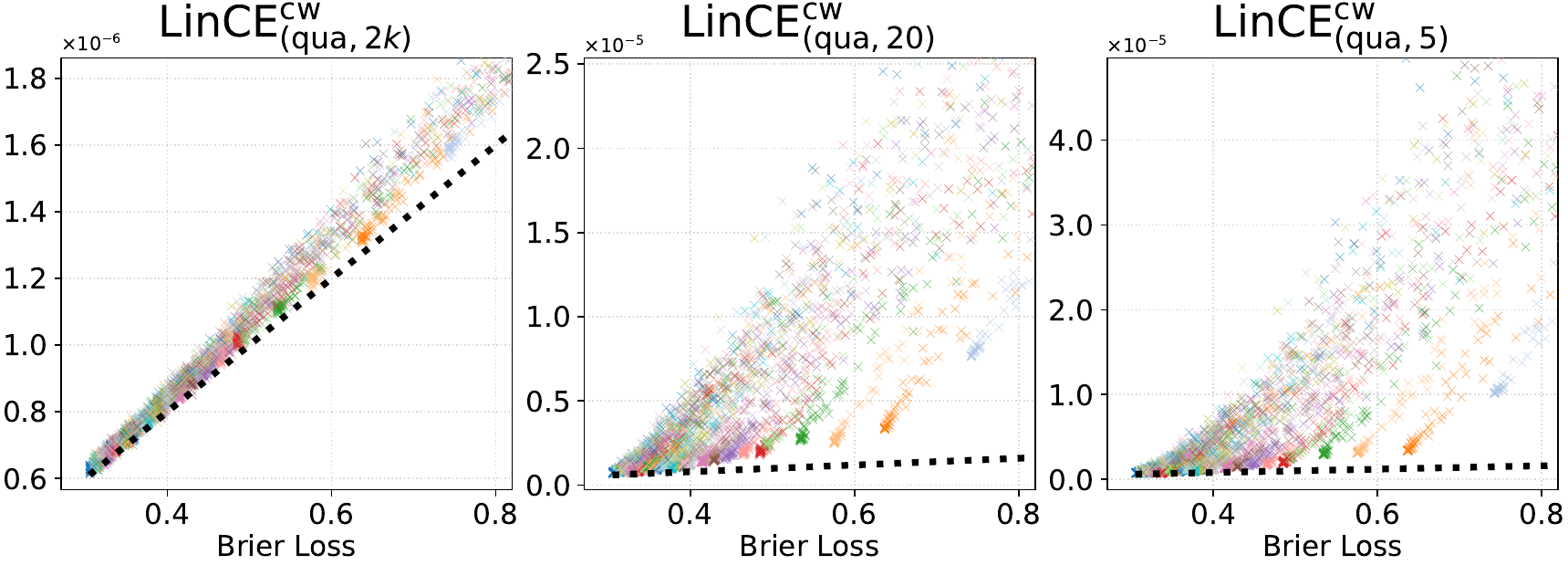}
        \caption{Within-model, classwise calibration.}
        \label{fig:estimate-loss-within-classwise}
    \end{subfigure}

    \vspace{0.8em}

    \begin{subfigure}[t]{0.49\linewidth}
        \centering
        \includegraphics[width=\linewidth]{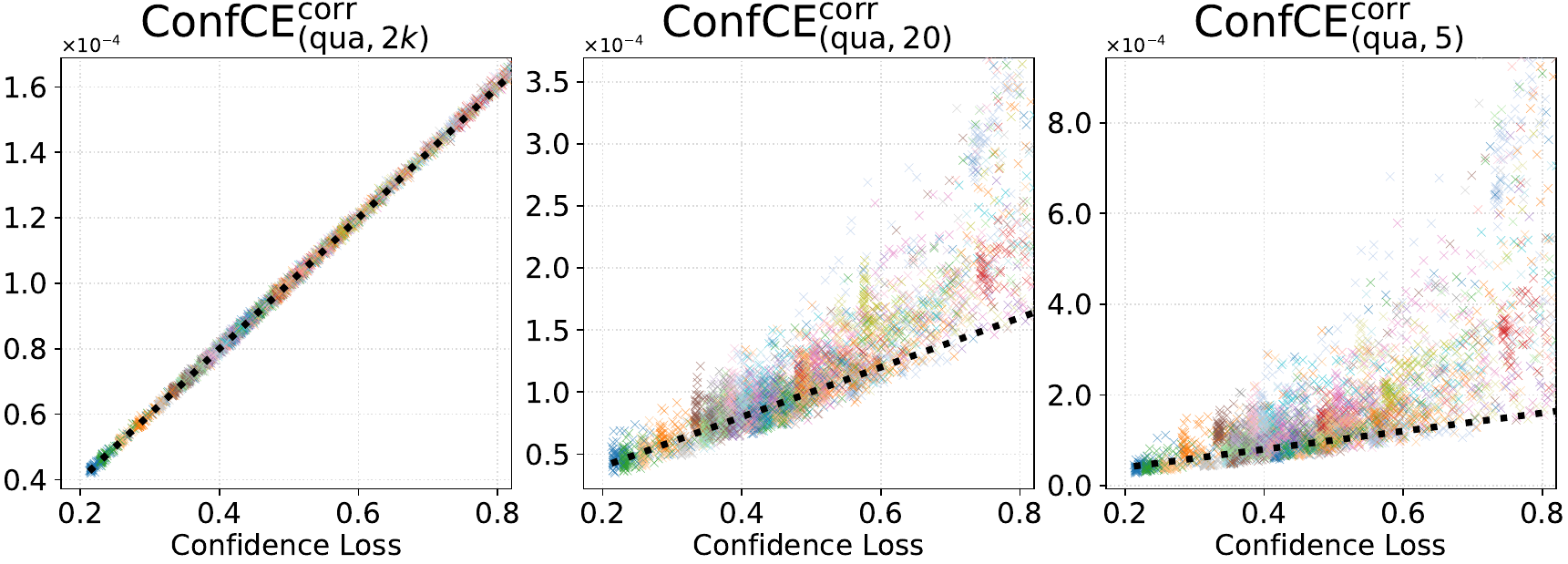}
        \caption{Cross-model, confidence calibration.}
        \label{fig:estimate-loss-cross-conf}
    \end{subfigure}
    \hfill
    \begin{subfigure}[t]{0.49\linewidth}
        \centering
        \includegraphics[width=\linewidth]{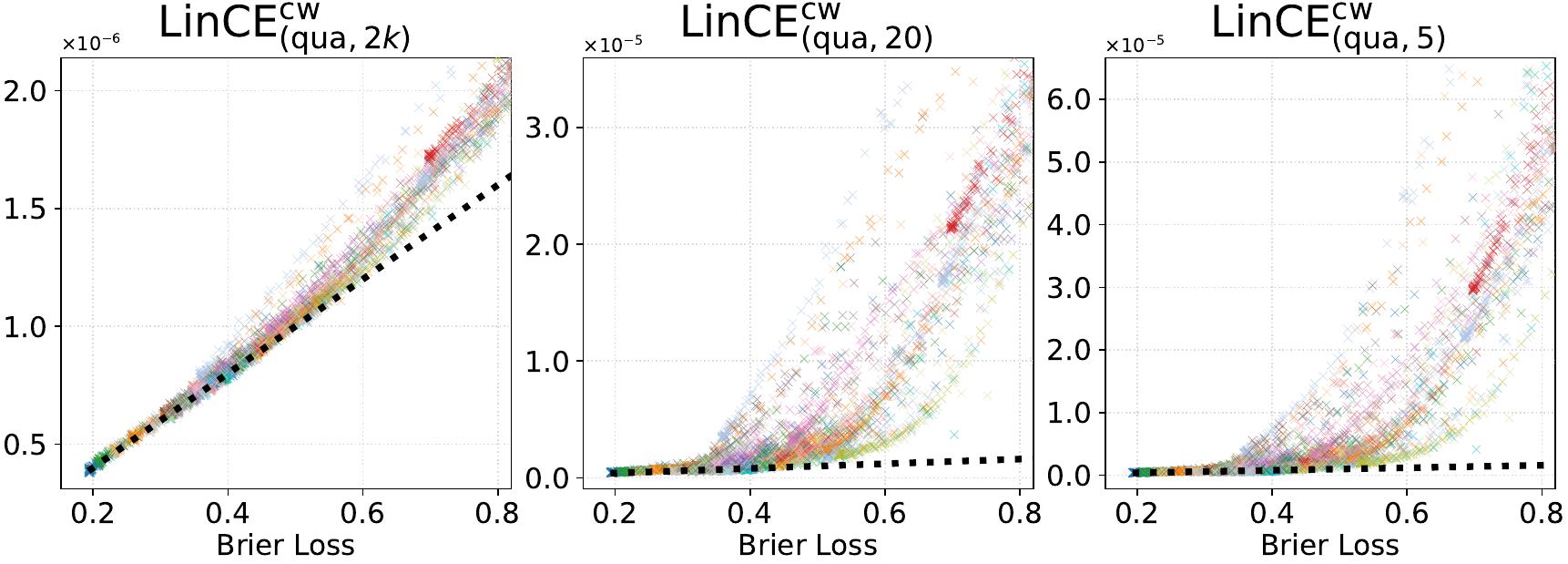}
        \caption{Cross-model, classwise calibration.}
        \label{fig:estimate-loss-cross-classwise}
    \end{subfigure}
    \caption{Assumption validation plots deferred from the main text. In each panel, the dotted line shows the calibration error predicted by our theory for a calibrated predictor as a function of the corresponding loss. After temperature scaling on the validation set, the within-model and cross-model checkpoints stay close to this line, supporting the approximately calibrated regime assumed in the empirical analysis.}
    \label{fig:estimate-loss-appendix}
\end{figure}

This section collects supplementary empirical plots deferred from the main
text. In particular, \Cref{fig:estimate-loss-appendix} provides the
assumption-validation plots showing that, after temperature scaling, the
evaluated predictors lie close to the calibrated-predictor loss lines for both
confidence and classwise calibration.

\subsection{Dominance Preservation and Binning Robustness: Omitted Figures}
\label{apdx: dominance preservation}

\begin{figure}[ht]
    \centering
    \begin{subfigure}[t]{0.49\linewidth}
        \centering
        \includegraphics[width=\linewidth]{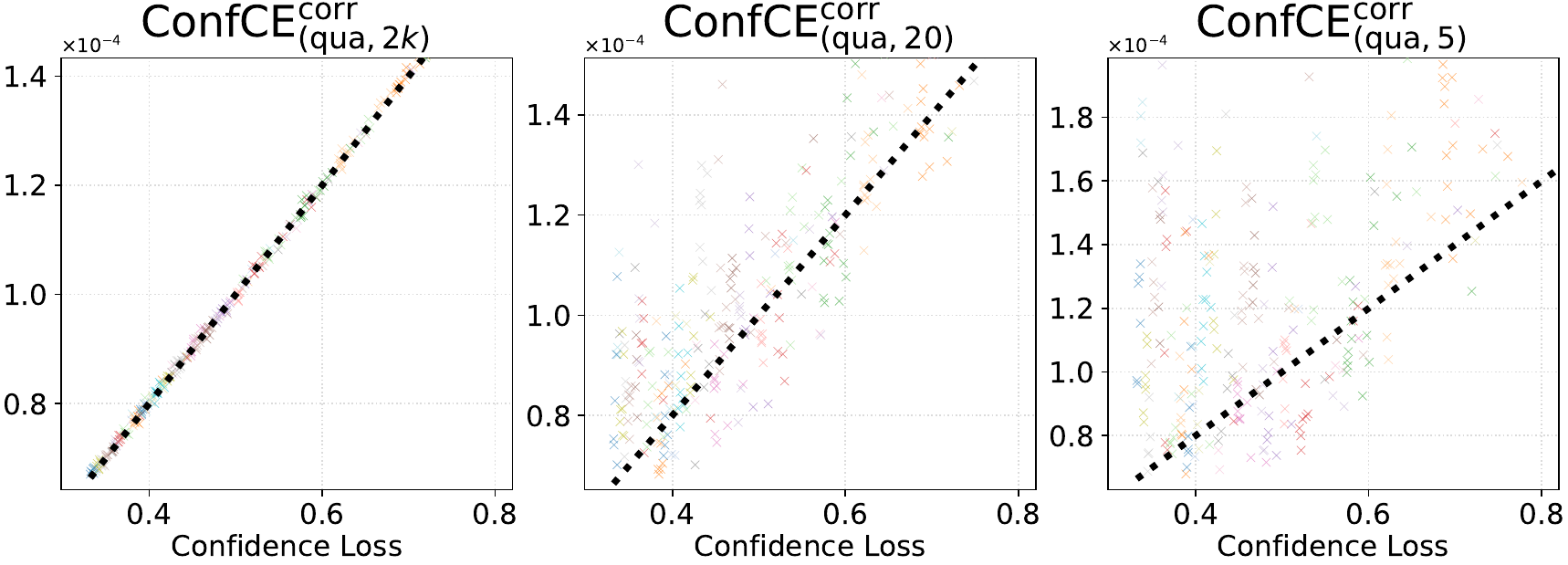}
        \caption{$\mathrm{ConfCE}^{\mathrm{corr}}_{\mathsf B}$ on the maximum dominance set.}
        \label{fig:dominance-a-app}
    \end{subfigure}
    \hfill
    \begin{subfigure}[t]{0.49\linewidth}
        \centering
        \includegraphics[width=\linewidth]{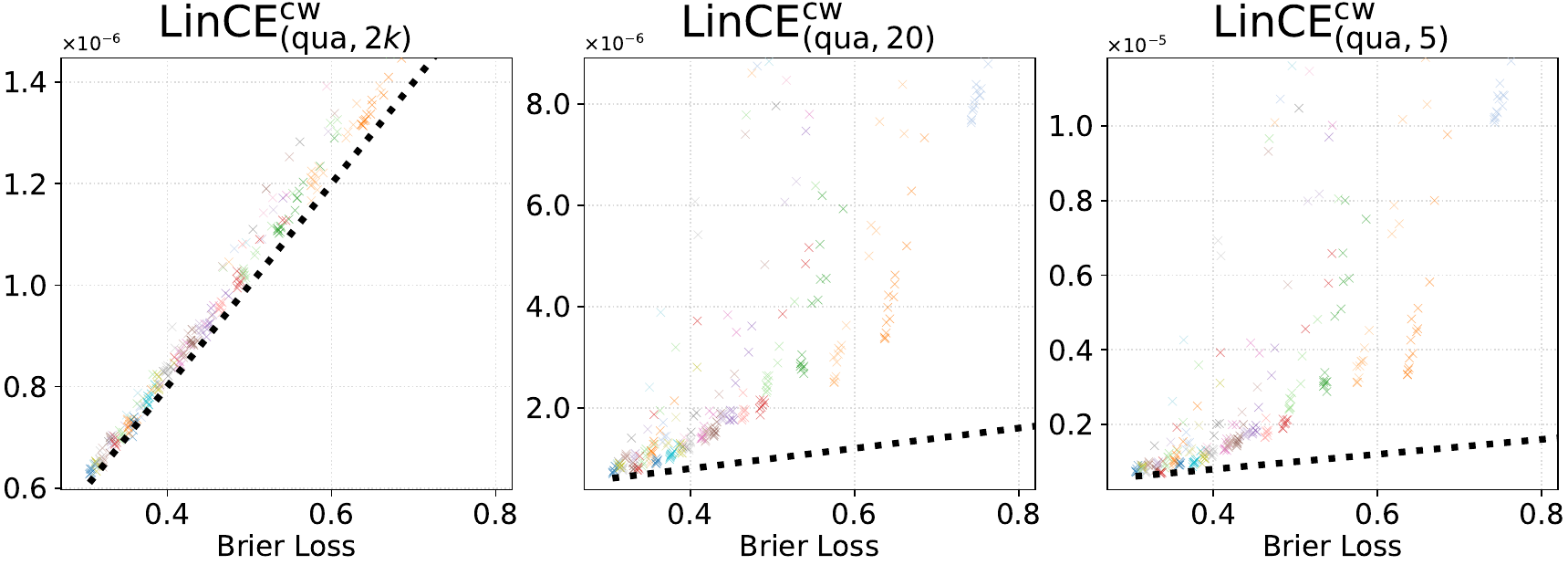}
        \caption{$\mathrm{LinCE}^{\mathrm{cw}}_{\mathsf B}$ on the maximum dominance set.}
        \label{fig:dominance-b-app}
    \end{subfigure}
    \caption{Empirical validation of dominance preservation on the within-model maximum dominance set with the same setting as \Cref{fig:dominance-ab}.}
    \label{fig:dominance-ab-app}
    \vspace{-3mm}
\end{figure}

\begin{figure}[htbp]
    \centering
    \begin{subfigure}[t]{0.49\linewidth}
        \centering
        \includegraphics[width=\linewidth]{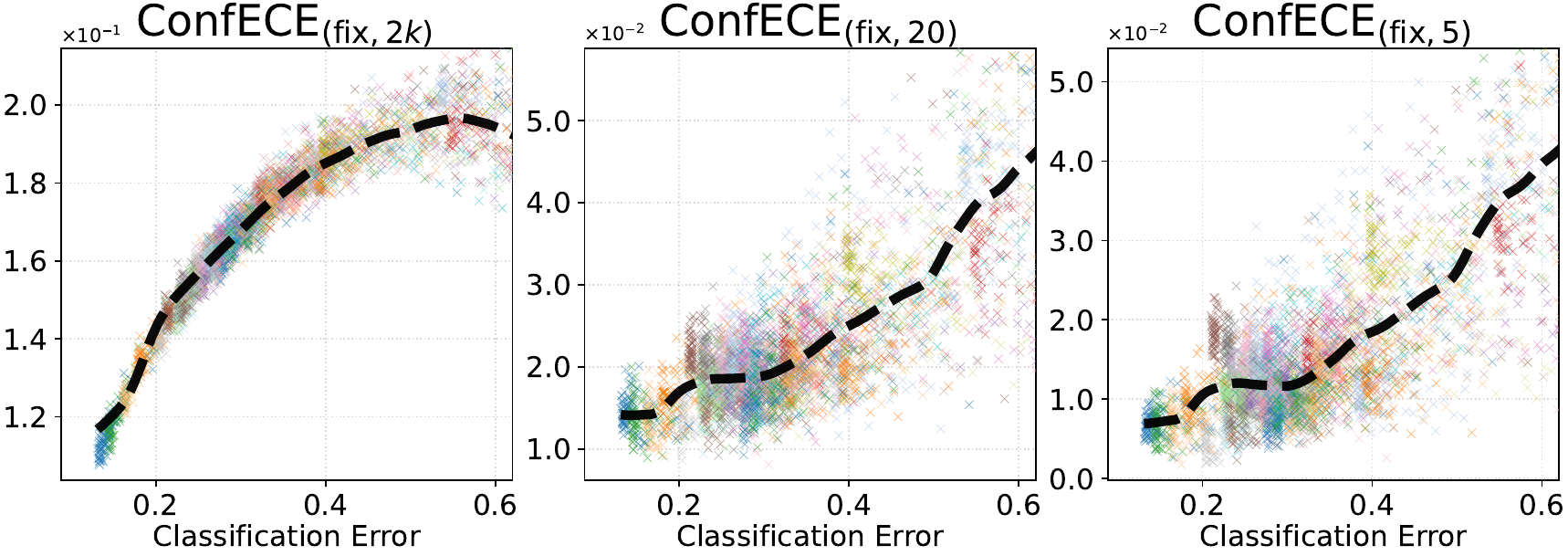}
        \caption{Non-truthful $\mathrm{ConfECE}_{\mathsf B}$.}
        \label{fig:trend-a-app}
    \end{subfigure}
    \hfill
    \begin{subfigure}[t]{0.49\linewidth}
        \centering
        \includegraphics[width=\linewidth]{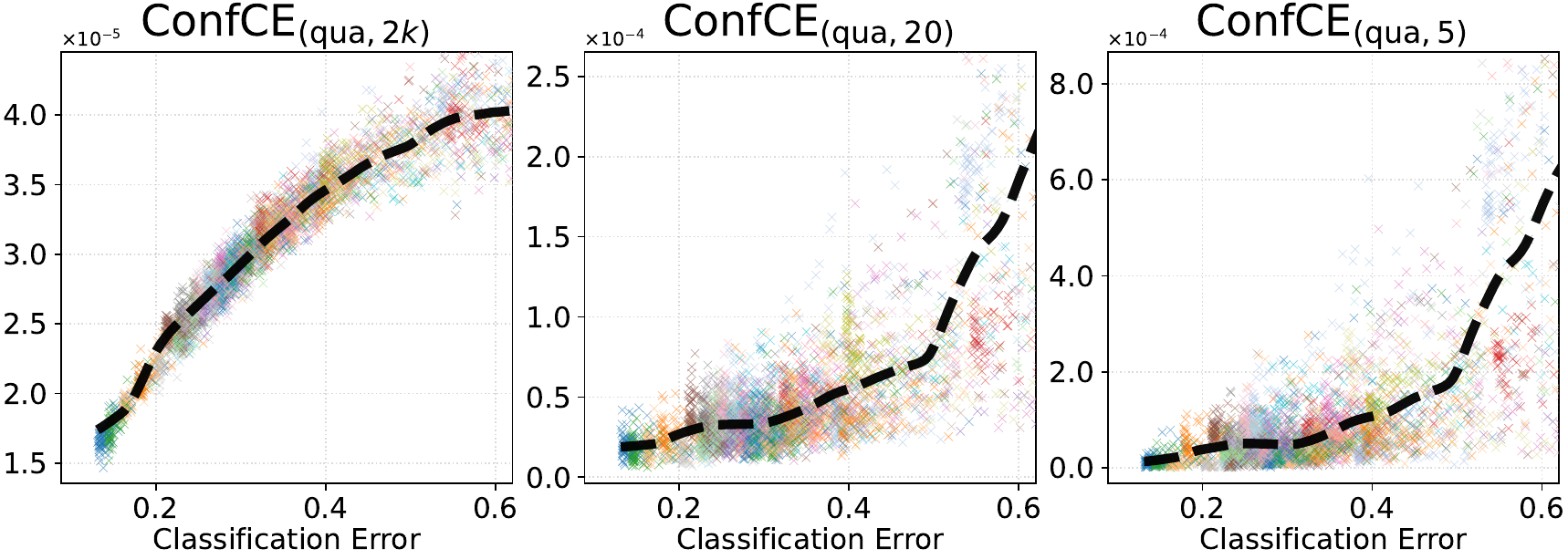}
        \caption{Non-truthful $\mathrm{ConfCE}_{\mathsf B}$.}
        \label{fig:trend-b-app}
    \end{subfigure}
\vspace{0.8em}
    \begin{subfigure}[t]{0.49\linewidth}
        \centering
        \includegraphics[width=\linewidth]{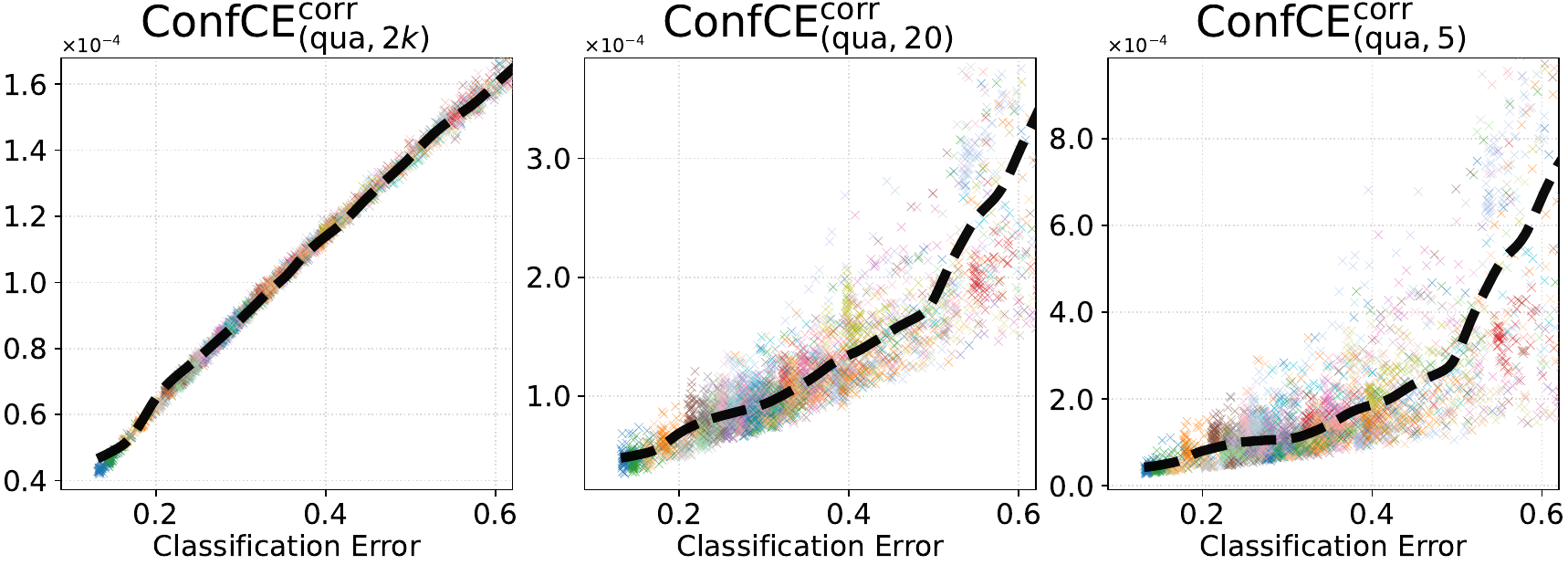}
        \caption{Truthful $\mathrm{ConfCE}^{\mathrm{corr}}_{\mathsf B}$.}
        \label{fig:trend-c-app}
    \end{subfigure}
    \hfill
    \begin{subfigure}[t]{0.49\linewidth}
        \centering
        \includegraphics[width=\linewidth]{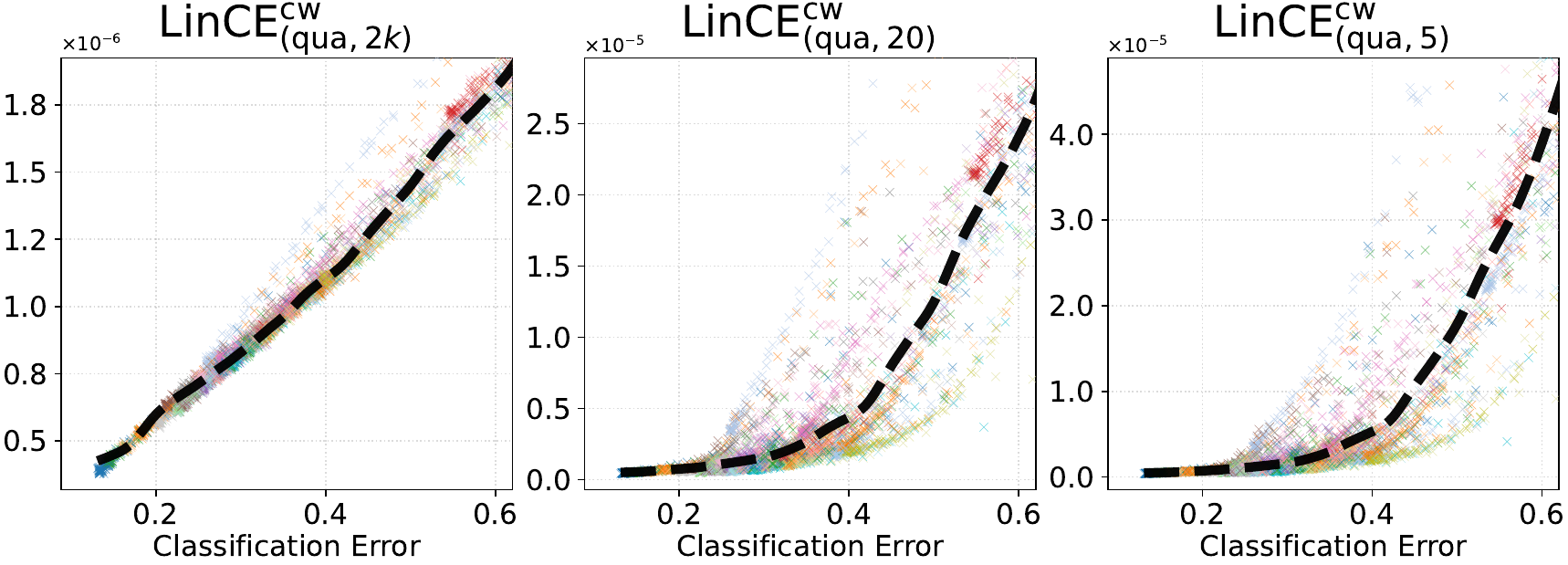}
        \caption{Truthful $\mathrm{LinCE}^{\mathrm{cw}}$.}
        \label{fig:trend-d-app}
    \end{subfigure}
    \vspace{-3mm}
    \caption{Binning Robustness on the cross-model traces with the same setting as \Cref{fig:trend-abcd}.}
    \label{fig:trend-abcd-app}
\end{figure}

% Generated by scripts/generate_trend_spearman_table.py
% Requires \usepackage{booktabs}, \usepackage{multirow}, and \usepackage{graphicx}.
\begin{table}[h]
\centering
\scriptsize
\resizebox{\linewidth}{!}{%
\begin{tabular}{lllll}
\toprule
Calibration Error & Item 1 & Item 2 & Within-Model Spearman & Cross-Model Spearman \\
\midrule
\multirow{6}{*}{$\mathrm{ConfECE}$} & Classification Error & $\mathrm{ConfECE}_{(\mathrm{fix}, 2k)}$ & 0.972 $\pm$ 0.002 & 0.974 $\pm$ 0.002 \\
 & Classification Error & $\mathrm{ConfECE}_{(\mathrm{fix}, 20)}$ & -0.009 $\pm$ 0.033 & 0.654 $\pm$ 0.018 \\
 & Classification Error & $\mathrm{ConfECE}_{(\mathrm{fix}, 5)}$ & -0.277 $\pm$ 0.030 & 0.604 $\pm$ 0.020 \\
 & $\mathrm{ConfECE}_{(\mathrm{fix}, 2k)}$ & $\mathrm{ConfECE}_{(\mathrm{fix}, 20)}$ & -0.006 $\pm$ 0.033 & 0.672 $\pm$ 0.018 \\
 & $\mathrm{ConfECE}_{(\mathrm{fix}, 2k)}$ & $\mathrm{ConfECE}_{(\mathrm{fix}, 5)}$ & -0.285 $\pm$ 0.030 & 0.624 $\pm$ 0.020 \\
 & $\mathrm{ConfECE}_{(\mathrm{fix}, 20)}$ & $\mathrm{ConfECE}_{(\mathrm{fix}, 5)}$ & 0.660 $\pm$ 0.019 & 0.841 $\pm$ 0.009 \\
\midrule
\multirow{6}{*}{$\mathrm{ConfCE}$} & Classification Error & $\mathrm{ConfCE}_{(\mathrm{qua}, 2k)}$ & 0.966 $\pm$ 0.002 & 0.976 $\pm$ 0.002 \\
 & Classification Error & $\mathrm{ConfCE}_{(\mathrm{qua}, 20)}$ & 0.045 $\pm$ 0.033 & 0.639 $\pm$ 0.019 \\
 & Classification Error & $\mathrm{ConfCE}_{(\mathrm{qua}, 5)}$ & -0.121 $\pm$ 0.032 & 0.582 $\pm$ 0.021 \\
 & $\mathrm{ConfCE}_{(\mathrm{qua}, 2k)}$ & $\mathrm{ConfCE}_{(\mathrm{qua}, 20)}$ & 0.070 $\pm$ 0.033 & 0.662 $\pm$ 0.018 \\
 & $\mathrm{ConfCE}_{(\mathrm{qua}, 2k)}$ & $\mathrm{ConfCE}_{(\mathrm{qua}, 5)}$ & -0.115 $\pm$ 0.032 & 0.608 $\pm$ 0.020 \\
 & $\mathrm{ConfCE}_{(\mathrm{qua}, 20)}$ & $\mathrm{ConfCE}_{(\mathrm{qua}, 5)}$ & 0.652 $\pm$ 0.019 & 0.828 $\pm$ 0.010 \\
\midrule
\multirow{6}{*}{$\mathrm{ConfCE}^{\mathrm{corr}}$} & Classification Error & $\mathrm{ConfCE}^{\mathrm{corr}}_{(\mathrm{qua}, 2k)}$ & 0.998 $\pm$ 0.000 & 0.999 $\pm$ 0.000 \\
 & Classification Error & $\mathrm{ConfCE}^{\mathrm{corr}}_{(\mathrm{qua}, 20)}$ & 0.691 $\pm$ 0.017 & 0.891 $\pm$ 0.007 \\
 & Classification Error & $\mathrm{ConfCE}^{\mathrm{corr}}_{(\mathrm{qua}, 5)}$ & 0.272 $\pm$ 0.031 & 0.753 $\pm$ 0.014 \\
 & $\mathrm{ConfCE}^{\mathrm{corr}}_{(\mathrm{qua}, 2k)}$ & $\mathrm{ConfCE}^{\mathrm{corr}}_{(\mathrm{qua}, 20)}$ & 0.695 $\pm$ 0.017 & 0.895 $\pm$ 0.006 \\
 & $\mathrm{ConfCE}^{\mathrm{corr}}_{(\mathrm{qua}, 2k)}$ & $\mathrm{ConfCE}^{\mathrm{corr}}_{(\mathrm{qua}, 5)}$ & 0.272 $\pm$ 0.031 & 0.760 $\pm$ 0.014 \\
 & $\mathrm{ConfCE}^{\mathrm{corr}}_{(\mathrm{qua}, 20)}$ & $\mathrm{ConfCE}^{\mathrm{corr}}_{(\mathrm{qua}, 5)}$ & 0.642 $\pm$ 0.020 & 0.884 $\pm$ 0.007 \\
\midrule
\multirow{6}{*}{$\mathrm{LinCE}^{\mathrm{cw}}$} & Classification Error & $\mathrm{LinCE}^{\mathrm{cw}}_{(\mathrm{qua}, 2k)}$ & 0.994 $\pm$ 0.000 & 0.992 $\pm$ 0.000 \\
 & Classification Error & $\mathrm{LinCE}^{\mathrm{cw}}_{(\mathrm{qua}, 20)}$ & 0.887 $\pm$ 0.007 & 0.875 $\pm$ 0.008 \\
 & Classification Error & $\mathrm{LinCE}^{\mathrm{cw}}_{(\mathrm{qua}, 5)}$ & 0.884 $\pm$ 0.007 & 0.873 $\pm$ 0.008 \\
 & $\mathrm{LinCE}^{\mathrm{cw}}_{(\mathrm{qua}, 2k)}$ & $\mathrm{LinCE}^{\mathrm{cw}}_{(\mathrm{qua}, 20)}$ & 0.899 $\pm$ 0.006 & 0.910 $\pm$ 0.006 \\
 & $\mathrm{LinCE}^{\mathrm{cw}}_{(\mathrm{qua}, 2k)}$ & $\mathrm{LinCE}^{\mathrm{cw}}_{(\mathrm{qua}, 5)}$ & 0.895 $\pm$ 0.007 & 0.907 $\pm$ 0.006 \\
 & $\mathrm{LinCE}^{\mathrm{cw}}_{(\mathrm{qua}, 20)}$ & $\mathrm{LinCE}^{\mathrm{cw}}_{(\mathrm{qua}, 5)}$ & 0.997 $\pm$ 0.000 & 0.997 $\pm$ 0.000 \\
\bottomrule
\end{tabular}
}
\caption{Spearman rank correlations on the test split for the calibration-error comparisons in our binning-robustness analysis. For each calibration-error family, the table reports the correlation between classification error and each calibration error, as well as the pairwise correlations among calibration errors computed with different bin choices. The last two columns compare the within-model MobileNetV3-Small-1.0 setting and the cross-model setting. The reported $\pm$ values denote $1.96$ standard deviations for confidence estimation. Truthful calibration errors remain stable across binning choices, whereas non-truthful calibration errors are substantially more sensitive to the choice of bin size. In particular, truthful errors display better rank correlation among errors computed with different binning sizes.}
\label{tab:trend-spearman}
\end{table}

\section{Experimental Details and Reproducibility}
\label{apdx:exp-repro}

All paper experiments use \texttt{CIFAR-100} distributed through
\texttt{torchvision}. The code keeps the official 50{,}000-image training split
intact and deterministically partitions the official 10{,}000-image test split
into 5{,}000 validation and 5{,}000 test examples using random seed 42. The
validation split is used only for post-hoc scaling; all reported figures and
tables are evaluated on the test split.

In the experiments, we fine-tune nine ImageNet-pretrained \texttt{timm} models:
\texttt{mobilenetv3\_small\_050}, \texttt{mobilenetv3\_small\_075},
\texttt{mobilenetv3\_small\_100}, \texttt{resnet10t}, \texttt{resnet18},
\texttt{resnet34}, \texttt{resnet50}, \texttt{resnet101}, and
\texttt{resnet152}. For \texttt{CIFAR-100}, all runs use resized
224\(\times\)224 inputs, ImageNet normalization, SGD with momentum 0.9,
learning rate \(10^{-3}\), weight decay \(5\times 10^{-4}\), cosine learning
rate schedule, batch size 256, 100 epochs, and mixed-precision training. The
training transform uses resize, random crop with padding 4, and random
horizontal flip; the evaluation transform uses deterministic resize. %Model
%suffixes of the form \texttt{\_\_trainpct<pct>\_seed<seed>} denote deterministic
%training subsets generated by a hashed permutation of training indices.

The within-model experiments use 40 traces from
\texttt{mobilenetv3\_small\_100}: the full-data run plus subset runs from
2.5\% to 97.5\% in increments of 2.5\%. The cross-model
experiments use 45 traces: for each of the nine architectures above, we use the
full-data run and subset runs at 20\%, 40\%, 60\%, and 80\%.

Every checkpoint is evaluated after temperature scaling. For each epoch, the temperature parameter is fit
on validation logits by minimizing validation cross-entropy with L-BFGS and is
then applied to test logits. The paper
presets use all available epochs except warm-up epochs (filtered by classification error) when generating the figures and tables.

All training runs were executed on the same machine using a single GPU. The
machine has dual Intel Xeon Platinum 8470Q CPUs (104 physical cores / 208
threads total), one NVIDIA RTX PRO 6000 Blackwell GPU with about 96\,GB device
memory, and 1.0\,TiB system RAM. The software stack uses PyTorch,
\texttt{torchvision}, and \texttt{timm}.

%%%%%%%%%%%%%%%%%%%%%%%%%%%%%%%%%%%%%%%%%%%%%%%%%%%%%%%%%%%%

\end{document}